\documentclass[10pt,twocolumn]{article}
\makeatletter
\providecommand{\Description}[1]{}
\makeatother
\usepackage[letterpaper,margin=0.75in]{geometry}
\usepackage{times}
\usepackage{microtype}

\usepackage{graphicx}
\usepackage{booktabs}
\usepackage{amsmath}
\usepackage{amssymb}
\usepackage{mathtools}
\usepackage{cuted}
\usepackage[numbers,sort&compress]{natbib}
\usepackage[ruled,vlined,linesnumbered]{algorithm2e}
\usepackage[hidelinks]{hyperref}

\SetAlFnt{\small}
\SetAlCapFnt{\small}
\SetAlCapNameFnt{\small}
\SetAlCapHSkip{0pt}

\begin{document}

\twocolumn[
  \begin{@twocolumnfalse}
    \begin{center}
      {\LARGE\bfseries Tired Actor: Fatigue-Informed Character Control\par}
      \vspace{0.75em}
      {\large
      Shengyuan Zhang$^1$,
      Xinpeng Liu$^{1,2}$,
      Muchun Niu$^1$,
      Yulong Chen$^3$,\\
      Lizhuang Ma$^1$,
      Yue Gao$^1$,
      Cewu Lu$^1$,
      Yong-Lu Li$^{1,2}$\textsuperscript{*}
      \par}
      \vspace{0.55em}
      {\normalsize
      $^1$Shanghai Jiao Tong University \quad
      $^2$Shanghai Innovation Institute \quad
      $^3$Harbin Institute of Technology
      \par}
      \vspace{0.55em}
      \vspace{0.35em}
    \end{center}
    \vspace{0.8em}
  \end{@twocolumnfalse}
]

\begin{abstract}
Replicating human behavior with physics simulation has been a long-expected
goal in character animation. Existing efforts have achieved impressive
performance in imitating a wide span of general motions. However, most existing
efforts could still suffer from \textbf{unnatural movements} due to the lack of
biomechanical and physiological priors. Given this, we project our sights to
advances in behavioral energetics, which demonstrate how energy use shapes
human movements. In contrast, current character controllers typically assume
the character is equipped with infinite energy over time. Inspired by these, we
propose to adopt \textbf{fatigue} as a proxy of the finite energy limit, inject
it into general character animation, and thoroughly investigate how fatigue
introduces new characteristics to physics-based character control. Leveraging
the Three-Compartment Controller (3CC) model, we managed to obtain a policy for
general motion imitation under different fatigue statuses. Furthermore,
extensive analyses are conducted to demonstrate how fatigue could influence the
naturalness, scalability, and robustness of character animation.
\textit{Our code will be made public.}
\end{abstract}


\section{Introduction}

\begin{figure*}[!t]
    \centering
    \includegraphics[width=\linewidth]{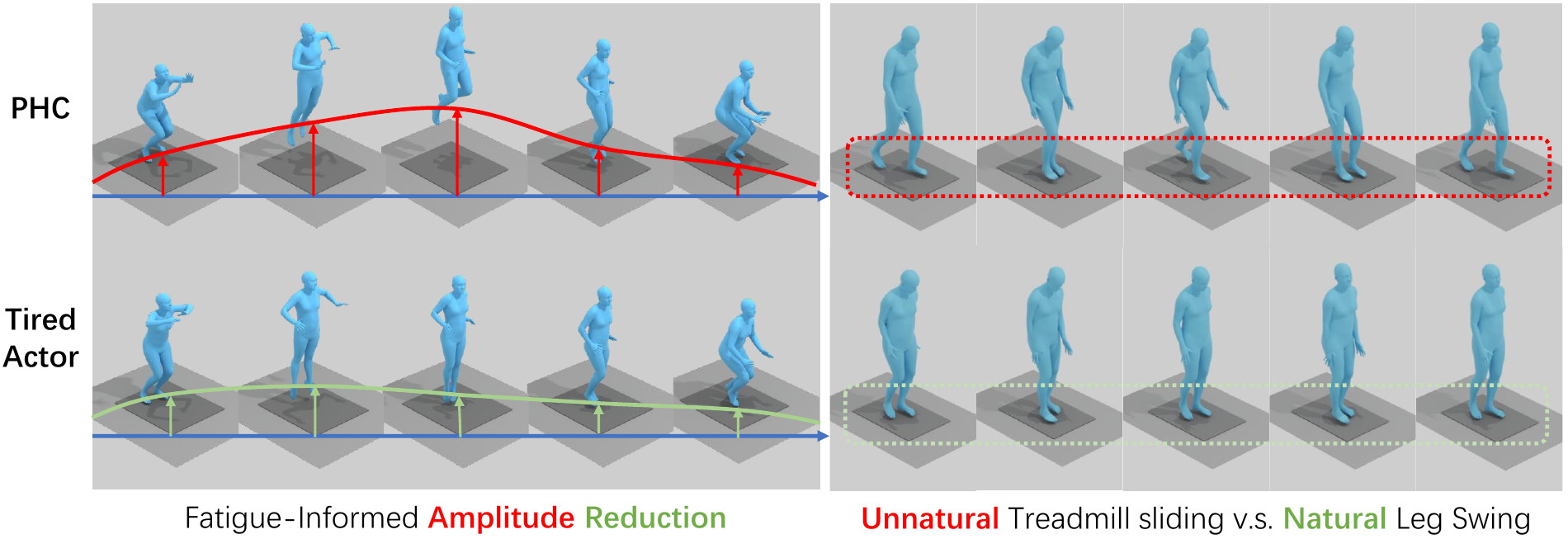}
    \vspace{-10px}
    \caption{Current controllers could still be unnatural, like replicating treadmill paces on the ground. Inspired by behavioral energetics, we introduce Tired Actor, a fatigue-informed character controller with interesting behaviors. As shown, our Tired Actor learns to cut corners by reducing action amplitudes. It also produces a reasonable and natural single-leg swinging motion instead of foot sliding on the ground, which is an interesting side effect. }
    \Description{Current controllers could still be unnatural, like replicating treadmill paces on the ground. Inspired by behavioral energetics, we introduce Tired Actor, a fatigue-informed character controller with interesting behaviors. As shown, our Tired Actor learns to cut corners by reducing action amplitudes. It also produces a reasonable and natural single-leg swinging motion instead of foot sliding on the ground, which is an interesting side effect. }
    \label{fig:cutting-corner}
    \vspace{-10px}
\end{figure*}

Physically controlling human-like characters to produce natural motions has been a long-pursued goal for the vision and graphics community, given its great potential in character animation~\cite{peng2022ase,zhu2023NCP}, virtual reality~\cite{winkler2022questsim,luo2024real}, and robotics~\cite{he2024hover}.
Tremendous progress has been achieved in reproducing human motions with simulated characters~\cite{luo2023perpetual,truong2024pdp}.
Moreover, deployments on humanoid robots are also made possible with impressive agility~\cite{he2025asap} and expressiveness~\cite{cheng2024expressive}.

Despite the impressive advances, existing efforts still produce \textit{unnatural movements under certain circumstances}, as shown in Fig.~\ref{fig:cutting-corner}, since most are purely data-driven with limited biomechanical and physiological priors.
Given this, we project our sight to an emerging research field termed behavioral energetics~\cite{behavioral}, which investigates how energy use shapes human movements. 
Research demonstrates how energy consumption influences the gaits and speed during locomotion~\cite{ralston1958energy,brooks2005basics}, gait transition~\cite{diedrich1995change,mercier1994energy}, cycling cadence~\cite{brisswalter2000energetically,riveros2025silico}, and navigation in dynamic contexts~\cite{brown2021unified}.
These efforts reveal that natural human movement patterns are tightly bound with energy optima.
Therefore, introducing energy sensing or its proxies to character control becomes a promising way to improve the naturalness.
In contrast, most efforts assume the character is equipped with \textit{infinite energy}.
The assumption's downside has not been noticed much, which is a side effect of the typical task formulation, where more attention is paid to short motion sequences.
The tracked motion sequences end before the cumulative tiredness could make an impact.
However, with the controllers becoming more and more satisfying, high hopes could be placed on the character control policies for lifelike movements with advanced applications like motion generation and stylization.
In this way, the human-likeness of characters could be a critical goal, and the infinitely energetic character is lower than expected.

Inspired by these, we propose to leverage fatigue as a proxy of energy-awareness and incorporate it with state-of-the-art character controllers, expecting to maintain the motion coverage while improving the naturalness of the character with behavioral energetics clues, as in Fig.~\ref{fig:cutting-corner}.
A Three-Compartment Controller (3CC) model~\cite{xia2008theoretical} adapted for torque-based motion control~\cite{cheema2023discovering} is adopted to represent the human fatigue mechanism, where the effect of fatigue is simulated via modifying the maximum applicable torque.
In detail, we first collect the maximal applicable torques by rolling out existing policies. 
Then, a fatigue-informed policy, Tired Actor, is trained concerning the collected maximal torques.
Moreover, we conduct thorough experiments on how fatigue reshapes the behaviors of the controlled character.
Specifically, we focus on three aspects.
First, different fatigue states could result in different motion states not in the original data distribution and encourage the policy to produce novel behaviors.
Relatedly, we are also curious about how the fatigue-extended data distribution might help the policy generalize to unseen behaviors.
Finally, inspired by the relationship between fatigue and compensation in human motion, we are also interested in whether fatigue-informed policy is equipped with proper compensation.

In summary, our main contribution is twofold:
First, inspired by behavioral energetics insights, we incorporate fatigue with a character policy termed Tired Actor, improving its naturalness while maintaining the motion coverage.
Second, we conduct thorough experiments to investigate how the fatigue mechanism could influence the diversity, generalization, and compensation.
\section{Related Works}

\subsection{Physics-Based Character Control}
Physics-based character control has been an active research field.
Early advances focused on tracking controllers for specific and limited motions~\cite{de2010feature,geijtenbeek2013flexible,liu2010sampling}.
With the development of deep Reinforcement Learning (RL) algorithms, success was achieved for imitating motion sequences with manually designed rewards~\cite{liu2017learning,liu2018learning,peng2018deepmimic}.
To reduce the reliance on heavy manual engineering, GAIL~\cite{ho2016generative} is introduced for human motion imitation~\cite{merel2017learning,peng2021amp}.
However, the generalization could still be limited.
To this end, with residual force control~\cite{yuan2020residual}, \citet{luo2021dynamics} managed to replicate over 97\% motion sequences in AMASS~\cite{mahmood2019amass}, achieving noticeable generalization ability.
\citet{luo2023perpetual} eliminated the supernatural residual force while maintaining the generalization ability. 
\citet{truong2024pdp} leveraged diffusion policy~\cite{chi2023diffusion} for better recovery ability.

With the well-developed motion-tracking controllers, controllers for more advanced tasks have become a new focus.
\citet{peng2022ase} proposed to learn skill embeddings for character controllers, providing an easy-to-use representation for new downstream tasks.
\citet{xu2023composite} learned decoupled skill representations for different body parts.
Towards better modeling of human skills, VAEs~\cite{yao2022controlvae,luo2024universal,tessler2024maskedmimic}, VQ-VAEs~\cite{zhu2023NCP,yao2024moconvq}, and diffusion models~\cite{truong2024pdp} are adopted to regulate the skill space.
Different high-level control signals, including text prompts~\cite{juravsky2022padl,kumar2023words,ren2023insactor,juravsky2024superpadl,tevet2024closd}, VR controller signals~\cite{winkler2022questsim,luo2024real}, and high-level action semantics~\cite{tessler2023calm,dou2022case} are also incorporated to character control.
Emerging efforts started to pay more attention to the interaction ability of the controlled actor.
\citet{luo2022embodied} leveraged a simulated character with scene-awareness for accurate 3D pose reconstruction.
A series of works aimed at controlling characters to produce feasible movements in different scenes~\cite{hassan2023synthesizing,lee2023questenvsim,xiao2023unified}, even producing highly agile parkour movements~\cite{xu2025parc}.
Moreover, controlling characters to interact with objects also gained attention.
Early efforts focused on graspable objects~\cite{braun2024physically,luo2024omnigrasp}, while recent advances grew the capability in whole-body manipulation~\cite{gao2024coohoi,xu2024humanvla,yu2025skillmimic,xu2025intermimic} and sport skills~\cite{wang2024strategy,yu2025skillmimic}.

Despite the impressive advances, most existing efforts were built on a purely data-driven basis, with limited exploration of biomechanical and physiological priors.
Unnatural movements could be produced under certain circumstances, as shown in Fig.~\ref{fig:cutting-corner}.
Given this, we resort to behavioral energetics for physiological prior.

\subsection{Behavioral Energetics}
Traditional views of biomechanics take the energetic cost as the result of human mechanics~\cite{donelan2004mechanical,gottschall2003energy,arellano2011effects}. 
Instead, behavioral energetics, as an emerging field, suggests that energetic cost plays an important role in shaping human movements.
Research has shown that humans (also other animals) tend to move with minimal energy use~\cite{atzler1927arbeitsphysiologische,elftman1966biomechanics,alexander1996optima}. 
Calories per unit distance are optimized concerning different walking speeds by humans~\cite{molen1972graphic,ralston1958energy}.
Gait parameters like step frequency and width are also energetically optimal~\cite{bertram2001multiple,holt1991predicting}.
The energetic optimum also appears when runners select a pace for sustainable running~\cite{rathkey2017people,steudel2009optimal,willcockson2012reconsidering}.
Some research also suggests the optima could be developed and refined via training~\cite{cher2015minimum}.
The causal effects between energy consumption and human movements are also observed in more complicated contexts.
Humans switch between walking and running at a speed, which becomes energetically more expensive to walk than to run under treadmill settings~\cite{diedrich1995change,hreljac1993preferred}.
In more natural settings, humans select among walking, running, and resting with dynamic speed adaptation to minimize the energy usage concerning the distance requirements and time limits~\cite{long2013walking}.
New evidence shows energy optimality could even explain how humans select path trajectories for turning and obstacle avoidance~\cite{brown2021unified,daniels2023human}.
Given these advances, we identify behavioral energetics as a promising inspiration for natural character control and introduce human fatigue as a proxy for energy sensing.

\subsection{Human Fatigue Modeling}
\citet{giat1993musculotendon,giat1996model} developed fatigue models for the quadriceps from fatigue-related metabolic parameters during functional electrical stimulation and recovery, while Ding et al.~\cite{ding2000predictive} introduced a four-parameter model to predict musculoskeletal fatigue.
Despite their realism, they were heavily based on detailed muscle activation patterns, limiting their application.
To this end, a motor unit-based fatigue was proposed~\cite{liu2002dynamical}, where the fatigue is estimated via three muscle states: activated, fatigued, and resting.
Based on this, the Three-Compartment Controller (3CC) model~\cite{xia2008theoretical} was introduced for dynamic load conditions.
\citet{cheema2023discovering} further extended it for torque-based scenarios.
There were also efforts in developing a data-driven model for human fatigue~\cite{kider2011data}, where a multi-modal capturing technique was developed.
Even less effort was paid to fatigue-aware character control.
Early efforts were made on human body parts, either for lower-body motion control~\cite{komura2000creating} or single-arm motion control~\cite{cheema2020predicting}.
\citet{feng2023musclevae} developed a fatigue-aware motion controller for muscle-actuated characters.
Concurrently, \citet{cheema2023discovering} enhanced AMP with fatigue.
However, both were limited in the scale of applicable motion sequences.
Also, the influence that the fatigue mechanism could impose has not been structurally investigated.

\section{Method}

\subsection{Fatigue-Informed Character Control Preliminaries}
\label{sec:control-algo}

We formulate our character controller $\pi(a^t|s^t_{task}, s_t^{self})$ w.r.t. the Markov Decision Process $\mathcal{M}=\langle \mathcal{S}, \mathcal{T}, \mathcal{A}, \mathcal{R}, \gamma \rangle$, composed of states $\mathcal{S}$ governed by the transition dynamics $\mathcal{T}$, available action space $\mathcal{A}$, reward function $\mathcal{R}$, and discount factor $\gamma$.
At each timestep $t$, given the current state $s^t$ and target state $s^{task}$, the control policy $\pi$ produces action $a^t\in \mathcal{A}$.
With transition dynamics $\mathcal{T}$, time proceeds, producing the state of the next timestep $s^{t+1}$.
Reward $r$ is then computed following the reward function $r^t=\mathcal{R}(s^{task}, s^{t+1}, s^t, a^t)$.
The training goal is to maximize the reward expectation $\mathbf{E}(\sum_{t=1}^T \gamma^{t-1}r^t)$.
Tired Actor, our trained controller, is designed to combine PHC~\citep{luo2023perpetual} and \cite{cheema2023discovering} to maintain the motion coverage while injecting the fatigue mechanism.

\noindent{\bf States $\mathcal{S}$.} 
A 24-joint SMPL~\citep{loper2023smpl} humanoid is adopted with full-zero body shape parameters.
At each time step, the state $s^t$ could be decoupled into two separate components: the self-state $s_{self}^t$ and the task state $s_{task}^t$. The self-state $s_{self}^t$ contains character proprioception information $s_{self}^t=\{h_{root}^t, x^t, \dot{x}^t, \theta^t, \dot{\theta}^t, M_F^t\} \in R^{427}$.
$h_{root}^t \in R$ represents the root height.
$x \in R^{69}, \dot{x} \in R^{72}$ are the positions and linear velocities of different body keypoints, canonicalized in the egocentric coordinate.
Therefore, the position of the root joint is eliminated from $x^t$.
$\theta \in R^{144}, \dot{\theta} \in R^{72}$ are the orientations and angular velocities of different body links, also canonicalized in the egocentric coordinate.
Finally, $M_F^t \in [0, 100]^{69}$ indicates the fatigue level at timestep $t$, which will be covered later.
The task state is defined as: 
\begin{equation}
\begin{aligned}
s_{\mathrm{task}}^t = \{&
\hat{x}^{t+1} - x^t,\;
\dot{\hat{x}}^{t+1} - \dot{x}^t,\;
\hat{\theta}^{t+1} - \theta^t, \\
&
\dot{\hat{\theta}}^{t+1} - \dot{\theta}^t,\;
\dot{\hat{\theta}}^{t+1},\;
\hat{x}^{t+1}
\}
\in \mathbb{R}^{576}.
\end{aligned}
\end{equation}
where $-$ indicates the rotation difference for $\theta,\dot{\theta}$.
Quantities with $\hat{\dot}$ are the corresponding next-step states in the reference motion.
All the quantities are in the egocentric coordinate.

\noindent{\bf Action $\mathcal{A}$.}
\label{sec:action}
All joints but the root (pelvis) are actuated with proportional derivative (PD) controllers.
The action $a^t$ is composed of two components as $a^t=\{\mu^t, \beta^t\} \in R^{70}$, where $\mu^t$ is the PD target, and $\beta^t$ is a stiffness and damping multiplier following \citep{cheema2023discovering}, which is expected to tune the tension of the whole body.
In practice, we also find that this decoupled action design makes it easier for the policy to adapt to different fatigue levels.
Without introducing the fatigue mechanism, the raw applied torque $\tau_{raw}^t$ could be computed as 
\begin{equation}
    \label{eq:tau}
    \tau_{raw}^t=\beta^t(k_p \circ (\mu^t-q^t) - k_d \circ \dot{q}^t),
\end{equation}
where $k_p,k_d$ are the stiffness and damping factors.

\noindent{\bf Transition Dynamics $\mathcal{T}$.}
We adopt IsaacGym~\citep{makoviychuk2021isaac} for rigid body dynamics simulation.
The transition dynamics are handled by the 3CC model~\citep{liu2002dynamical,xia2008theoretical,cheema2023discovering}, which assumes motor units are either active ($M_A$), fatigued ($M_F$), or resting ($M_R$), represented in the percentage of maximum voluntary torques. 
Initially, all motor units are resting ($M_R=100\%$).
Resting units become activated once a target load ($TL$) appears, resulting in increased $M_A$.
Activated units could then be fatigued over time, with descending $M_A$ and increasing $M_F$.
Meanwhile, the fatigued units could also regain their power over time, resulting in $M_F$ decreasing and $M_R$ increasing.
The transition dynamics of fatigue could be described as
\begin{equation}
    \begin{aligned}
        \frac{\partial M_A}{\partial t} &= C(TL) - F\cdot M_A, \\
        \frac{\partial M_F}{\partial t} &= F \cdot M_A - R_r \cdot M_F, \\
        \frac{\partial M_R}{\partial t} &= -C(TL) + R_r \cdot M_F, \\
    \end{aligned}
\end{equation}
\begin{equation}
    \begin{aligned}
        R_r &= \begin{cases}
            R\cdot r, \ \rm{if} \ M_A \geq TL, \\
            R, \ \rm{else}, 
        \end{cases}\\
    \end{aligned}
\end{equation}
\begin{equation}
    \begin{aligned}
        C(TL) &= \begin{cases}
            L_R\cdot (TL - M_A), \ \rm{if} \ M_A \geq TL, \\
            L_D\cdot (TL - M_A), \ \rm{if} \ TL - M_R< M_A < TL, \\
            L_D\cdot M_R \ \rm{else}.
        \end{cases}
    \end{aligned}
\end{equation}the fatigue and recovery rates.
$1 - \frac{R}{F}$ also indicates the upper bound of $M_F$.
$r$, as an additional rest recovery multiplier for intermittent tasks~\citep{looft2018modification}. 
$r > 1$ indicates that the recovery is faster when not all active units are necessary to achieve the target load (TL).
$C(TL)$ is a bounded proportional controller, producing the required TL while dynamically changing the proportion of $M_A$ and $M_F$.
$C(TL)$ is characterized by muscle development factor $L_D$ and relaxation factor $L_R$, preventing $M_A$ and $M_F$ from instantaneous changes, thus simulating the muscle activation dynamics~\citep{thelen2003adjustment,winters1995improved}.
With this, the applied torque in Sec.~\ref{sec:action} is further modified according to the fatigue state as 
\begin{equation}
    \begin{aligned}
        TL &= \frac{\tau_{raw}^t}{\tau_{max}}, RC = 1 - M_F, \\
        \tau^t &= clip(\tau_{raw}^t, -RC \cdot \tau_{max}, RC \cdot \tau_{max}).
    \end{aligned}
\end{equation}
Here, $\tau_{max}$ represents the maximal exertable torque, and $\tau^t$ is the final applied torque.

\noindent{\bf Rewards.}
We define the reward function as $r_t = 0.5r^t_{task} + 0.5r_{amp}^t + r_t^{p} + r_t^{f}$.
In detail, the task reward is defined as $r_{task}^t=\omega_1e^{-100\|\hat{x}^t-x^t\|} + \omega_2e^{-0.1\|\hat{\dot{x}}^t-\dot{x}^t\|} + \omega_3e^{-10\|\hat{\theta}^t-\theta^t\|} + \omega_4e^{-0.1\|\hat{\dot{\theta}}^t-\dot{\theta}^t\|}$, 
where the translation and linear velocities of all joints, and the orientation and angular velocities of all links are considered.
For reward $r_{amp}^t$, we follow AMP~\citep{peng2021amp} for the observations, loss formulation, and gradient penalty design.
For power reward $r_{p}^t$, we follow \citep{luo2023perpetual} as $r_{p}=-0.0005|\dot{\theta}^t\tau^t|^2$, encouraging low-energy movements which tend to look more natural~\citep{yu2018learning}.
We further introduce a fatigue reward $r_{f}^t=0.01M_F^t$ for explicit regulation on cumulative fatigue minimization.

\subsection{Tired Actor: Fatigue-Informed Character Controller}
\label{sec:training-algo}
\begin{figure}[!h]
    \centering
    \includegraphics[width=\linewidth]{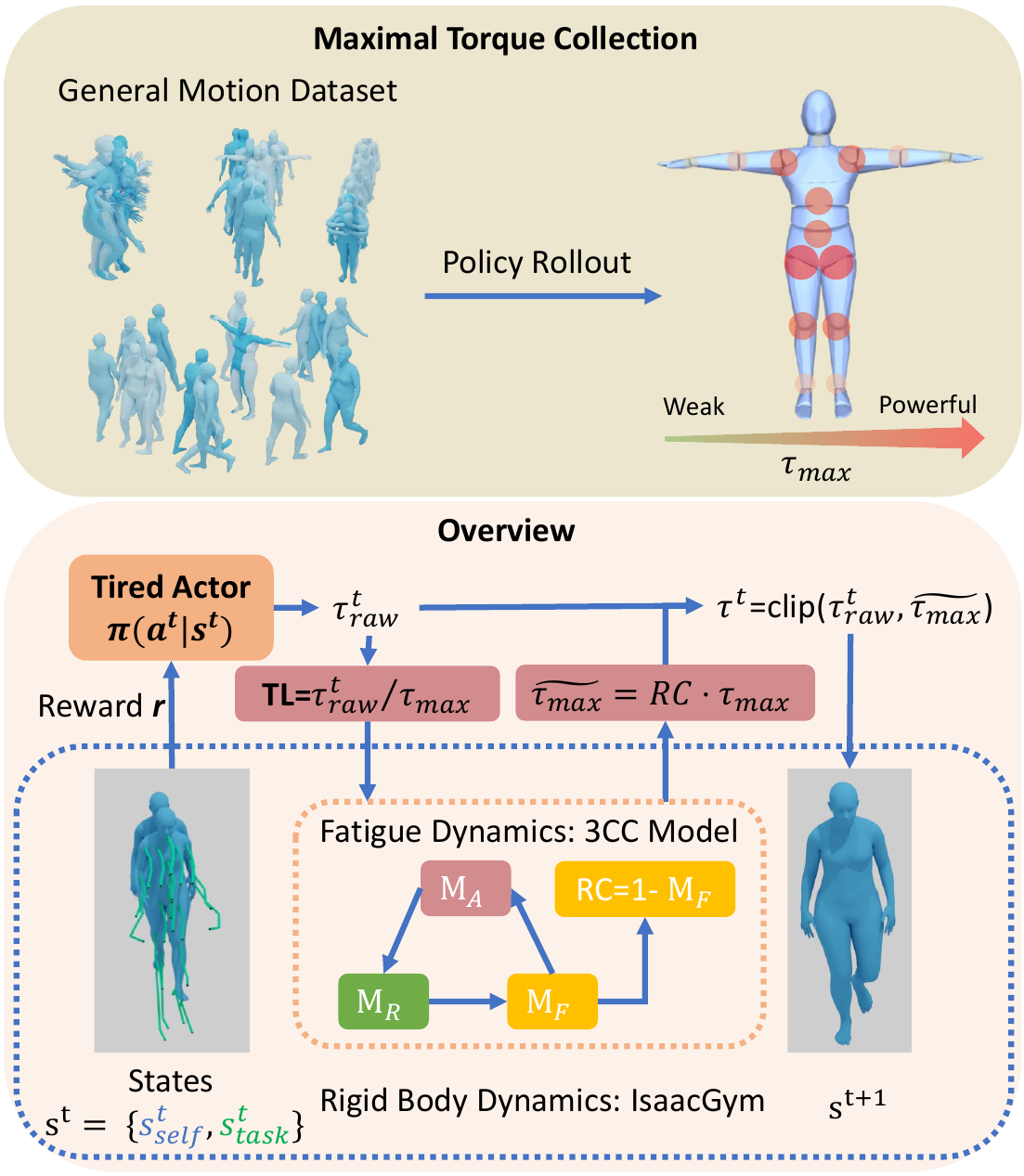}
    \caption{The general pipeline of Tired Actor. We first collect the maximal torque limits from AMASS, a large-scale general motion dataset. Then, we train the Tired Actor with fatigue-awareness.}
    \Description{The general pipeline of Tired Actor. We first collect the maximal torque limits from AMASS, a large-scale general motion dataset. Then, we train the Tired Actor with fatigue-awareness.}
    \label{fig:pipeline}
    \vspace{-10px}
\end{figure}

Next, we introduce our fatigue-informed controller: Tired Actor.
It is simply designed as $\pi(a^t|s^t)=\mathcal{N}(\mu(s_t), \sigma(s_t))$, where $\mu(\cdot),\sigma(\cdot)$, and the discriminator for $r_{amp}$ are all MLPs.
We first collect maximal torque limits by rolling out existing motion controllers.
Then, the Tired Actor is trained with reweighted fatigue initialization, hard negative mining, and fall-recovery finetuning.
The general pipeline is shown in Fig.~\ref{fig:pipeline}.

\noindent{\bf Maximal Torque Collection. }
A key component for the fatigue-informed controller is the maximal torque $\tau_{max}$, which defines the maximal applicable torque.
Previous effort~\citep{cheema2023discovering} pre-trained per-sequence motion controllers to obtain per-action torque limits.
This might be preferable for single-action control.
Instead, aiming at making Tired Actor compatible with general motion sequences, we directly roll out PHC on AMASS~\citep{mahmood2019amass}, and record the maximal torques for each joint according to Eq.~\ref{eq:tau} with $\beta^t=1$.
Thus, the obtained torque limits could function more like a representation of general human motor capacity than an action-intensity representation.

\noindent{\bf Reweighted Fatigue Initialization.}
\label{sec:reweighted-init}
\begin{figure}[!h]
    \centering
    \includegraphics[width=.8\linewidth]{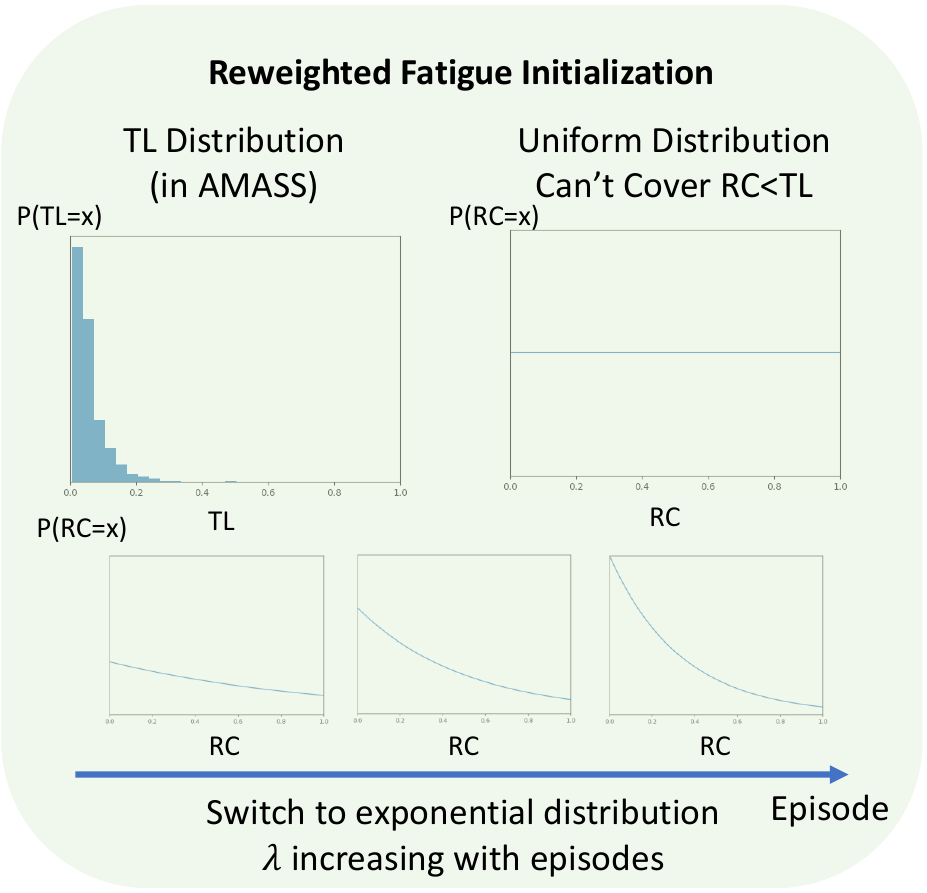}
    \caption{Reweighted fatigue initialization. }
    \Description{Reweighted fatigue initialization. }
    \label{fig:reweight}
    \vspace{-10px}
\end{figure}
With the collected torque limits, we could now train our fatigue-informed controller with PPO. 
The criterion network shares the same structure as the policy.
At the start of each episode, the initial fatigue states $M_F,M_A,M_R$, and fatigue parameters $F,R,r$ are randomly sampled.
However, this introduces new problems.
In \cite{cheema2023discovering}, $M_F,M_A,M_R$ was uniformly sampled.
However, our general-purpose torque limits could be much higher than single-action limits.
For most actions, $TL$ is less than $10\%$, meaning that even $M_F=90\%$ could make little difference in motion patterns.
Thus, the original uniform distribution could not cover the $RC < TL$ scenario well, which is important for learning fatigued movement patterns.
Therefore, we introduce reweighted fatigue initialization as shown in Fig.~\ref{fig:reweight}.
Instead of uniform distribution for $RC (=1-M_F)$, we first sample $RC$ with exponential distribution $Exp(\lambda_{ep})$ whose parameter $\lambda_{ep}$ increases with the number of episodes, denoted as 
\begin{equation}
    RC \sim \exp(\lambda_{ep}), M_F=1-\rm{clip}(RC, 0, 1).
\end{equation}
In this way, we make more explorations in the high-fatigue region, encouraging the policy to focus on fatigue-related motion patterns.
Meanwhile, the varying $\lambda_{ep}$ ensures a smooth learning procedure, preventing the model from failing in the early stage due to difficult, high-fatigue scenarios.

\noindent{\bf Hard Negative Mining.}
\label{sec:hard-neg}
Hard negative mining is a proven-successful technique in previous efforts~\cite{luo2021dynamics,luo2023perpetual}.
During training, we regularly evaluate our trained controller on the whole dataset and identify the hard sequences as the sequences in which our controller fails.
The sampling weights of these hard sequences are increased.
By failure, we follow the definition in \citep{luo2021dynamics,luo2023perpetual} as the state with mPJPE greater than 0.5 meters.

\noindent{\bf Fall-recovery Finetuning.}
In PHC, the fall-recovery ability is achieved by training an extra primitive for recovery and ensembling it with other primitives.
Instead, we directly fine-tune our controller with fall-recovery scenarios after achieving general motion imitation abilities.
We follow a simplified design of that in \citep{luo2023perpetual}.
At the beginning of randomly decided episodes, we simulate without applying any torques for a certain time step, making the character fall.
Then, we start training from the fallen states.
We did not adopt the fallen-state reward in \citep{luo2023perpetual} due to our limited controller capacity.
Even though we observe an increasing ability in hard sequence imitation and fall recovery with the finetuning procedure.

\section{Experiments}
\subsection{Implementation Details}
Tired Actor is trained on the same AMASS~\cite{mahmood2019amass} training split as PHC~\cite{luo2023perpetual}.
The actor and critic networks are seven-layer MLPs with the hidden size of [2048, 1536, 1024, 1024, 512, 512] and SiLU activation.
The AMP discriminator is a three-layer MLP with a hidden size of [1024, 512] and ReLU activation.
The policy is first trained for 30K episodes, then finetuned with fall-recovery for 20K episodes.
$\lambda_{ep}$ for the reweighted fatigue initialization is initially set as 1.
It gradually increases to 3 in the first 20K episodes, then remains fixed at 3.
NVIDIA's IsaacGym~\cite{makoviychuk2021isaac} is adopted for physics simulation.
The policy runs at 30Hz, while the simulation runs at 60Hz.
All experiments are conducted on a single NVIDIA RTX3090 GPU with 1,536 concurrent environments.
PPO is adopted for policy training, with a learning rate of 2e-5, discount factor $\gamma=0.99$.
For reward $r_{task}$, we set $\omega_1=0.5, \omega_2=0.1, \omega_3=0.3, \omega_4=0.1$.
During training, we conduct an evaluation every 5000 episodes to update the hard negative sampling weight.

\subsection{Fatigue-Informed Motion Imitation}
We first evaluate the performance of Tired Actor on motion imitation. 
Following PHC~\citep{luo2023perpetual}, we report the Success Rate.
When mPJPE$>0.5 m$, we deem the imitation as failed and terminate the imitation.
The global mPJPE-G and root-relative mPJPE-L are both reported in millimeters.
Also, the acceleration error and velocity errors are reported.
\begin{table*}[!t]
    \centering
    \resizebox{.95\linewidth}{!}{
    \begin{tabular}{lcccccccccc}
        \toprule
                    & \multicolumn{5}{c}{AMASS-Train} & \multicolumn{5}{c}{AMASS-Test} \\
        Method      & Success Rate (\%) & $mPJPE-G$ & $mPJPE-L$ & Acc. Error & Vel. Error & Success Rate & 
        $mPJPE-G$ & $mPJPE-L$ & Acc. Error & Vel. Error\\
        \hline
        PHC         & 98.9          & 37.5          & 26.9          & \textbf{3.3} & \textbf{4.9} & 96.4          & \textbf{47.4} & 30.9          & 6.8          & 9.1 \\
        Baseline    & 98.5          & \textbf{34.7} & 27.0          & \textbf{3.3} & \textbf{4.9} & 96.4          & 51.8          & 32.3          & 6.0          & 8.4 \\
        Tired Actor & \textbf{99.1} & 37.4          & \textbf{25.4} & 3.7          & 5.1          & \textbf{97.9} & 47.6          & \textbf{30.1} & \textbf{5.7} & \textbf{8.2} \\
        \bottomrule
    \end{tabular}}
    \vspace{10px}
    \caption{Results of motion imitation on AMASS.}
    \vspace{-10px}
    \label{tab:amass}
\end{table*}
\begin{table}[!t]
    \centering
    \resizebox{.9\linewidth}{!}{
    \begin{tabular}{lcccc}
    \toprule
              &\multicolumn{2}{c}{Baseline} & \multicolumn{2}{c}{Tired Actor} \\
    Parts     & mPJPE-G & mPJPE-L & mPJPE-G     & mPJPE-L \\
    \hline
    Pelvis    & 30.2          & -             & \textbf{28.6} & -             \\
    \hline
    Hips      & 32.3          & 9.1           & \textbf{30.2} & \textbf{8.8}  \\
    Knees     & \textbf{42.5} & \textbf{33.8} & 45.6          & 38.8          \\
    Ankles    & 60.2          & 53.2          & \textbf{54.4} & \textbf{51.9} \\
    \hline
    Shoulders & \textbf{30.1} & 19.5          & 31.4          & \textbf{18.5}  \\
    Elbows    & \textbf{31.4} & 28.5          & 33.8          & \textbf{26.0}  \\
    Wrists    & \textbf{35.1} & 36.7          & 37.7          & \textbf{31.9}  \\
    \bottomrule     
    \end{tabular}}
    \vspace{10px}
    \caption{Part-level mPJPE comparison on AMASS-Train with initial $M_F=M_A=0, M_R=100$.}
    \label{tab:parts}
    \vspace{-10px}
\end{table}
Besides PHC, we also train a baseline by eliminating the fatigue modeling of the Tired Actor.
The fatigue parameters of Tired Actor are $F=2, R=0.05, r=1$.
The initial fatigue state is set as $M_F=M_A=0, M_R=100$.
Quantitative results are shown in Tab.~\ref{tab:amass}.
On the AMASS train split, Tired Actor achieves the highest success rate.
The advantage of Tired Actor is also seen on mPJPE-L.
However, Tired Actor is noticeably worse than baseline on mPJPE-G, acceleration, and velocity imitation.
These indicate fatigue might encourage the sacrifice of the global pose following ability in exchange for the local pose imitation ability.
Also, the velocity and acceleration are giving way to the optimization of mPJPE-L.
For the unseen AMASS test split, Tired Actor surprisingly provides better generalization ability, surpassing the baseline on all metrics.
A more detailed analysis of per-part mPJPE is provided in Table~\ref{tab:parts}.
The tradeoff is not only involving global and local tracking, but also entangled with the lower-body and upper-body.
Noticeably, the Tired Actor is superior in following the root trajectory. 
The knee joint, as the key source of lower-body agility, trades its tracking precision for the precision of hips and ankles.
For the upper body, Tired Actor typically prefers local pose resemblance.

We further demonstrate some qualitative samples in Fig.~\ref{fig:fatigue-level}.
The Tired Actor learns to cut corners when imitating the reference motion by reducing action amplitudes and trailing behind. 
With the initial $M_F$ increasing, the model tends to put less effort and exhibits smaller action amplitudes.
We also compute the step length of the Tired Actor and baseline on 500 walking sequences, where the Tired Actor presents a noticeably lower step length of 0.385 m compared to the baseline with 0.407 m.
Interestingly, the cutting-corner behavior could improve the motion naturalness under certain scenarios, as shown in Fig.~\ref{fig:cutting-corner}. 
Instead of stubbornly imitating the treadmill walking motion on static ground like PHC, Tired Actor swings a single leg to balance between resemblance and naturalness. 

\subsection{Fatigue-Informed Behaviors}
\begin{table*}[!t]
    \centering
    \resizebox{\linewidth}{!}{
    \begin{tabular}{lcccccccccc}
        \toprule
                    & \multicolumn{5}{c}{AMASS-Train} & \multicolumn{5}{c}{AMASS-Test} \\
        Method      & Success Rate & mPJPE-G & mPJPE-L & Acc. Error & Vel. Error & Success Rate & mPJPE-G & mPJPE-L & Acc. Error & Vel. Error\\
        \hline
        Tired Actor ($M_F$=0)  & \textbf{99.1} & \textbf{37.4} & \textbf{25.4} & 3.7          & 5.1          & \textbf{97.9} & 47.6          & 30.1          & \textbf{5.7} & \textbf{8.2} \\
        Tired Actor ($M_F$=50) & 98.4 & \textbf{37.4} & 26.5          & \textbf{3.4} & \textbf{5.0} & 97.1          & \textbf{46.9} & \textbf{29.7} & 5.8          & 8.3 \\
        Tired Actor ($M_F$=80) & 98.0 & 38.0          & 27.2          & 3.5          & \textbf{5.0} & 95.7          & 47.7          & 30.1          & 5.8          & 8.3 \\
        Tired Actor ($M_F$=90) & 96.9 & 39.1          & 28.0          & 3.6          & 5.2          & 94.3          & 49.3          & 31.4          & 6.0          & 8.6 \\
        Tired Actor ($M_F$=95) & 91.1 & 44.0          & 31.1          & 4.7          & 6.1          & 89.3          & 52.5          & 33.6          & 6.7          & 9.3 \\
        \bottomrule
    \end{tabular}}
    \vspace{10px}
    \caption{Motion imitation with different initial $M_F$s.}
    \label{tab:mf}
    \vspace{-10px}
\end{table*}

\noindent{\bf Different Initial $M_F$.}  
Besides the qualitative demonstrations in Fig.~\ref{fig:fatigue-level}, quantitative results are reported in Tab.~\ref{tab:mf} with shared fatigue parameter $F=2, R=0.05, r=1$.
$M_F$ introduces a marginal performance drop until reaching about $90\%$, consistent with the fact that most movements only need less than $20\%$ maximum torque.
Also, not all the metrics reach the best level with the initial $M_F=0$. 
It might be easier to learn some motions with ``weaker'' characters.

\noindent{\bf Different Body Parts.} 
Fig.~\ref{fig:different-part} illustrates how fatigue of different parts could influence character motion.
When initializing the arms with $M_F=99\%$ and others with $M_F=0$, the arms only experience a short period of stiffness with a quick recovery.
This indicates that fatigue of distal segments could be easily handled.
We also initialize the legs with $M_F=99\%$ and others with zero $M_F=0$, compared to a character with whole-body initial $M_F=99\%$.
The former falls due to a strength disorder in different joints.
Instead, the latter balances by coordinating the weak but consistent joints, showing the critical role of consistency for fatigue compensation.

\noindent{\bf Left-Right Asymmetry. }
As shown in Fig.~\ref{fig:left-right}, we initialize half of the human body with $M_F=50\%$, and the other half with $M_F=0\%$ when imitating a circle-running motion. 
The fatigue parameters are set as $F=2, R=r=0$.
When the left (inner) half is fatigued, noticeable deviation is observed. 
With the fatigue accumulating, the character could even collapse. 
In contrast, the fatigue of the right (outer) half only introduces reduced action amplitude with marginal deviation. 
The asymmetric response to inner- and outer-half fatigue reflects the asymmetry of the underlying dynamic mechanism. 
Walking in circles might require the inner half of the body to exert higher torques to simultaneously maintain the gait and turn directions.

\noindent{\bf Spatial Decision-Making.}
Previous efforts~\citep{brown2021unified,daniels2023human} demonstrated the correlation between energetic optima and spatial decision-making procedures like path selections.
To identify whether similar behaviors exist for Tired Actor, we first distill Tired Actor into a student policy as $\pi_d(a^t|s^t_{task}, s_t^{self})$ following DAGGER~\citep{ross2011reduction}. 
Instead of the next-frame imitation target (Eq.~\ref{eq:task-state}) used as $s_t^{task}$ for the vanilla Tired Actor, $s_t^{task}$ is defined as: $s_{task}^t = \{kp_{task} - kp^t\}$, where $kp$ is the positions of a subset of human body keypoints, and $kp_{task}$ indicates a long-term target.
In this way, the student policy is capable of selecting its own path towards a target pose, rather than strictly following the given path.
To probe the spatial decision-making characteristics of Tired Actor, we put it in a simple path-following setting.
The path is designed as a polyline with one turning angle and two target root locations. 
Every three seconds, we switch to the next target root location. 
Each episode ends in six seconds.
As in Fig.~\ref{fig:spatial-decision}, Tired Actor avoids sharp turns and exerts lower accelerations towards the changing target, resulting in lower trajectory curvatures.
Also, we find that Tired Actor managed to follow the trajectory more precisely compared to the baseline.
However, when faced with rather sharp turns, Tired Actor could trade the target reaching timeliness for a smoother trajectory.

\noindent{\bf Different Fatigue Parameters $F,R,r$.}
We find that a high $F$ causes the character to fall quickly, substantially degrading the performance.
Quantitatively, keeping $R=0.05, r=1$ while increasing $F$ from 2 to 10 results in a $76.5\%$ success rate on AMASS-Test. 
If we then increase the recovery rate $R$ from 0.05 to 0.2, the success rate could increase to $93.2\%$.
However, the rest recovery multiplier $r$ makes less contribution compared to $F$ and $R$, and barely influences the success rate.
Qualitative demonstrations are included in the supplementary video.

\noindent{\bf Different body parts.}
In Fig.~\ref{fig:perturbation-part}, perturbations are applied to different body parts.
Perturbations applied to distal segments like hands are easily handled with little trajectory deviation, suggesting that disturbances in peripheral regions are more readily compensated for. 
In contrast, forces on central regions like the torso induce instability and falls. 
This pattern reflects the importance of maintaining core stability within the control policy.

\subsection{Fatigue-Informed Generalization}
\begin{table*}[!t]
    \centering
    \resizebox{\linewidth}{!}{
    \begin{tabular}{lccccccccc}
        \toprule
                    & \multicolumn{3}{c}{AMASS-Train} & \multicolumn{3}{c}{AMASS-Test} & \multicolumn{3}{c}{CIRCLE}\\
        Method      & Success Rate & mPJPE-G & mPJPE-L  & Success Rate & 
        mPJPE-G & mPJPE-L & Success Rate & mPJPE-G & mPJPE-L\\
        \hline
        Baseline (100 seq)        & 85.8          & \textbf{43.1} & \textbf{32.2} & \textbf{63.6} & \textbf{77.7} & 47.3          & 54.5          & \textbf{169.6} & 114.2 \\
        Tired Actor (100 seq)     & \textbf{87.5} & 50.1          & 33.1          & 61.4          & 81.7          & \textbf{46.9} & \textbf{61.4} & 172.7          & \textbf{111.9} \\
        \hline
        Baseline (1,000 seq)       & 97.2          & \textbf{35.3} & \textbf{28.2} & 88.6          & \textbf{51.1} & \textbf{34.9} & 66.6          & 164.9          & 113.3 \\
        Tired Actor (1000 seq)    & \textbf{98.4} & 40.6          & 29.5          & \textbf{90.0} & 55.3          & 36.1          & \textbf{67.9} & \textbf{163.4} & \textbf{109.1} \\         
        \hline
        Baseline                  & 98.5          & \textbf{34.7} & 27.0          & 96.4          & 51.8          & 32.3          & 70.2          & 163.9          & 109.2 \\
        Tired Actor               & \textbf{99.1} & 37.4          & \textbf{25.4} & \textbf{97.9} & \textbf{47.6} & \textbf{30.1} & \textbf{71.4} & \textbf{163.3} & \textbf{103.4} \\    
        \bottomrule
    \end{tabular}}
    \vspace{10px}
    \caption{Quantitative results for fatigue-informed generalization.}
    \vspace{-10px}
    \label{tab:generalization}
\end{table*}
The novel behaviors introduced by fatigue also extend the original distribution.
To this end, we train and evaluate variants of the baseline and the Tired Actor by limiting the training data scale to 100 and 1,000 sequences.
The evaluation is conducted on three datasets.
The AMASS train set (11,313 sequences) is adopted for seen in-domain evaluation.
The AMASS test set (140 sequences) is adopted for unseen in-domain evaluation.
Besides AMASS, CIRCLE~\citep{araujo2023circle} is adopted for unseen out-of-domain evaluation, given its 7,200 whole-body reaching motion sequences.
As shown in Tab.~\ref{tab:generalization}, Tired Actor performs consistently better than the non-fatigue baseline in success rate on the seen AMASS train set, with slightly worse mPJPE-G.
Also, the baseline is better at mPJPE-L with limited data scale, but the performance gap becomes smaller and smaller with the increasing training data scale.
For the unseen in-domain evaluation on the AMASS test set, the baseline generally provides better performance with small-scale training data, but is finally surpassed by Tired Actor with sufficient data.
When it comes to the unseen OOD evaluation on CIRCLE, both suffer from the domain gap.
However, it is noticeable that Tired Actor is consistently better with different training data scales, demonstrating its better generalization ability.

From the quantitative results, we can observe a common macro trend: Tired Actor is equipped with a better generalization to handle unseen motion sequences, and the gain effect increases with the data scale.
These trends coincide with our hypothesis that fatigue extends the data distribution.
With more training data, better extensions could be made.
We can also observe the trade-off of Tired Actor.
Typically, Tired Actor prioritized the success rate, then mPJPE-L, and finally mPJPE-G.
This manifests that by introducing fatigue, the controller is encouraged to keep balance and local pose resemblance, in exchange for less precise global pose tracking.
Finally, an interesting observation is the surprising performance when trained on only 100 sequences.
Even with very limited motion sequences, the controllers could provide competitive performance, indicating the inherent simplicity of human motor ability and inspiring future works on data-efficient character control.

\subsection{Fatigue-Informed Compensation}

We further investigate the Tired Actor's ability to compensate for external perturbations. 
We apply a 400-Newton force for 1s at a certain part's centroid, starting from the same time step. 
More details are available in the appendix.

\noindent{\bf Different Fatigue Levels.} 
In Fig.~\ref{fig:perturbation-fatigue}, we initialize actors with $MF=0$ and $MF=90$. 
The force is applied to the pelvis in the forward direction. 
As demonstrated, PHC falls and quickly recovers with unnatural motion.
In contrast, the Tired Actor with initial $M_F=0$ manages to resist the applied force and avoids falling.
The Tired Actor with initial $M_F=90$ falls with a more reasonable gesture.
These suggest that by introducing fatigue, the Tired Actor is enhanced in its compensation ability.

\noindent{\bf Different Force Directions.}
In Fig.~\ref{fig:perturbation-direction}, the external force is applied at the pelvis from different directions of the global velocity direction. 
Backward perturbation forces are the most destabilizing, causing falls and deviations that challenge the recovery process. 
Lateral perturbations induce moderate instability, whereas downward forces are absorbed more effectively, resulting in minor disruption. 
These findings indicate anisotropic sensitivity to directional disturbances. 

\subsection{Ablation Studies}
\noindent{\bf Hard-Negative Mining.} 
Tired Actor without hard-negative mining achieves 95.2\% success rate, 43.2 mPJPE-G, and 30.5 mPJPE-L on AMASS-Train. On AMASS-Test, it achieves a 92.3\% success rate, 52.5 mPJPE-G, and 35.1 mPJPE-L.
The performance degradation indicates the efficacy of focusing more on learning from hard sequences.

\noindent{\bf Fall-Recovery Finetuning.} 
Fig.~\ref{fig:getup} illustrates the Tired Actors with and without fall-recovery finetuning imitating a forward-walking motion with initial $M_F=95\%$. 
Tired Actor managed to avoid falling, though the gesture is twisted to some extent.
Without fall-recovery fine-tuning, the character fails to follow the reference motion and falls on the ground.
This reflects that the fall-recovery fine-tuning helps improve the robustness of the controller.

\section{Discussion}
Although Tired Actor achieves better naturalness while maintaining the wide motion coverage, some limitations remain. 
Compared to \citep{cheema2023discovering}, the emerging behavior patterns of Tired Actor are mostly limited to reduced action amplitude, while other expected motor strategies, like waiting, are rarely observed. 
We identify this as a result of our strict task setting of per-frame motion imitation. 
With the reference motion as a strong constraint, the Tired Actor struggles and falls rather than waiting for recovery from fatigue.
Therefore, extending Tired Actor to less-strict scenarios would be a promising goal.
Additionally, the generalization ability of Tired Actor, although impressive, remains limited for out-of-domain data.
Given the fatigue-informed data distribution extension, it would be interesting to exploit the generalization potential of motion control further.
Finally, though the Tired Actor is enhanced in natural compensation against external perturbations, unnatural gestures could still be observed. 
Also, the behavior after falling could collapse.
There are still substantial spaces for reducing the unnaturalness in character control.
\section{Conclusion}
Inspired by insights from behavioral energetics, we introduced Tired Actor, a character controller incorporated with fatigue modeling, which maintained general motion coverage while achieving better naturalness.
We further thoroughly investigated how the fatigue mechanism could influence the diversity, generalization, and robustness of character animation. We believe this work could open more possibilities in enhancing life-like character control.

Looking ahead, we envision that the principles established in this study could also inspire advancements in humanoid robotics, particularly in optimizing energy efficiency for extended operation and informing proactive joint maintenance strategies.

\bibliographystyle{ACM-Reference-Format}
\bibliography{sample-bibliography}

@String{Computing = "Computing" }

@String{Computer = "{IEEE} Computer" }

@String{Springer = "Springer-Verlag" }

@inproceedings{juravsky2022padl,author = {Juravsky, Jordan and Guo, Yunrong and Fidler, Sanja and Peng, Xue Bin},
title = {PADL: Language-Directed Physics-Based Character Control},
year = {2022},
isbn = {9781450394703},
publisher = {Association for Computing Machinery},
address = {New York, NY, USA},
url = {https://doi.org/10.1145/3550469.3555391},
doi = {10.1145/3550469.3555391},
booktitle = {SIGGRAPH Asia 2022 Conference Papers},
articleno = {19},
numpages = {9},
location = {Daegu, Republic of Korea},
series = {SA '22}
}

@inproceedings{tevet2024closd,
  author       = {Guy Tevet and
                  Sigal Raab and
                  Setareh Cohan and
                  Daniele Reda and
                  Zhengyi Luo and
                  Xue Bin Peng and
                  Amit Haim Bermano and
                  Michiel van de Panne},
  title        = {CLoSD: Closing the Loop between Simulation and Diffusion for multi-task
                  character control},
  booktitle    = {The Thirteenth International Conference on Learning Representations,
                  {ICLR} 2025, Singapore, April 24-28, 2025},
  publisher    = {OpenReview.net},
  year         = {2025},
  url          = {https://openreview.net/forum?id=pZISppZSTv},
  bibsource    = {dblp computer science bibliography, https://dblp.org}
}

@inproceedings{luo2024real,author = { Luo, Zhengyi and Cao, Jinkun and Khirodkar, Rawal and Winkler, Alexander and Huang, Jing and Kitani, Kris and Xu, Weipeng },
booktitle = { 2024 IEEE/CVF Conference on Computer Vision and Pattern Recognition (CVPR) },
title = {{ Real-Time Simulated Avatar from Head-Mounted Sensors }},
year = {2024},
volume = {},
ISSN = {},
pages = {571-581},
doi = {10.1109/CVPR52733.2024.00061},
url = {https://doi.ieeecomputersociety.org/10.1109/CVPR52733.2024.00061},
publisher = {IEEE Computer Society},
address = {Los Alamitos, CA, USA},
month =Jun
}

@article{giat1993musculotendon,
  title={A musculotendon model of the fatigue profiles of paralyzed quadriceps muscle under FES},
  author={Giat, Yohanan and Mizrahi, Joseph and Levy, Mark and others},
  journal={IEEE transactions on biomedical engineering},
  volume={40},
  number={7},
  pages={664--674},
  year={1993}
}

@article{giat1996model,
    author = {Giat, Yohanan and Mizrahi, Joseph and Levy, Mark},
    title = {A Model of Fatigue and Recovery in Paraplegic’s Quadriceps Muscle Subjected to Intermittent FES},
    journal = {Journal of Biomechanical Engineering},
    volume = {118},
    number = {3},
    pages = {357-366},
    year = {1996},
    month = {08},
    issn = {0148-0731},
    doi = {10.1115/1.2796018},
    url = {https://doi.org/10.1115/1.2796018},
    eprint = {https://asmedigitalcollection.asme.org/biomechanical/article-pdf/118/3/357/5766662/357\_1.pdf},
}

@article{liu2002dynamical,
  title={A dynamical model of muscle activation, fatigue, and recovery},
  author={Liu, Jing Z and Brown, Robert W and Yue, Guang H},
  journal={Biophysical journal},
  volume={82},
  number={5},
  pages={2344--2359},
  year={2002},
  publisher={Elsevier}
}

@article{ding2000predictive,
  title={A predictive model of fatigue in human skeletal muscles},
  author={Ding, Jun and Wexler, Anthony S and Binder-Macleod, Stuart A},
  journal={Journal of applied physiology},
  volume={89},
  number={4},
  pages={1322--1332},
  year={2000},
  publisher={American Physiological Society Bethesda, MD}
}

@misc{xu2025intermimic,
    title={InterMimic: Towards Universal Whole-Body Control for Physics-Based Human-Object Interactions},
    author={Sirui Xu and Hung Yu Ling and Yu-Xiong Wang and Liang-Yan Gui},
    year={2025},
    eprint={2502.20390},
    archivePrefix={arXiv},
    primaryClass={cs.CV}
}

@misc{yu2025skillmimic,
      title={SkillMimic-V2: Learning Robust and Generalizable Interaction Skills from Sparse and Noisy Demonstrations}, 
      author={Runyi Yu and Yinhuai Wang and Qihan Zhao and Hok Wai Tsui and Jingbo Wang and Ping Tan and Qifeng Chen},
      year={2025},
      eprint={2505.02094},
      archivePrefix={arXiv},
      primaryClass={cs.LG},
      url={https://arxiv.org/abs/2505.02094}, 
}

@inproceedings{wang2024strategy,
author = {Wang, Jiashun and Hodgins, Jessica and Won, Jungdam},
title = {Strategy and Skill Learning for Physics-based Table Tennis Animation},
year = {2024},
isbn = {9798400705250},
publisher = {Association for Computing Machinery},
address = {New York, NY, USA},
url = {https://doi.org/10.1145/3641519.3657437},
doi = {10.1145/3641519.3657437},
booktitle = {ACM SIGGRAPH 2024 Conference Papers},
articleno = {122},
numpages = {11},
location = {Denver, CO, USA},
series = {SIGGRAPH '24}
}

@article{thelen2003adjustment,
  title={Adjustment of muscle mechanics model parameters to simulate dynamic contractions in older adults},
  author={Thelen, Darryl G},
  journal={J. Biomech. Eng.},
  volume={125},
  number={1},
  pages={70--77},
  year={2003}
}

@article{winters1995improved,
  title={An improved muscle-reflex actuator for use in large-scale neuromusculoskeletal models},
  author={Winters, Jack M},
  journal={Annals of biomedical engineering},
  volume={23},
  pages={359--374},
  year={1995},
  publisher={Springer}
}

@article{looft2018modification,
  title={Modification of a three-compartment muscle fatigue model to predict peak torque decline during intermittent tasks},
  author={Looft, John M and Herkert, Nicole and Frey-Law, Laura},
  journal={Journal of biomechanics},
  volume={77},
  pages={16--25},
  year={2018},
  publisher={Elsevier}
}

@inproceedings{xu2024humanvla,
 author = {Xu, Xinyu and Zhang, Yizheng and Li, Yong-Lu and Han, Lei and Lu, Cewu},
 booktitle = {Advances in Neural Information Processing Systems},
 editor = {A. Globerson and L. Mackey and D. Belgrave and A. Fan and U. Paquet and J. Tomczak and C. Zhang},
 pages = {18633--18659},
 publisher = {Curran Associates, Inc.},
 title = {HumanVLA: Towards Vision-Language Directed Object Rearrangement by Physical Humanoid},
 url = {https://proceedings.neurips.cc/paper_files/paper/2024/file/215aeb07b5996c969c0123c3c6ee8f54-Paper-Conference.pdf},
 volume = {37},
 year = {2024}
}

@article{gao2024coohoi,
  title={Coohoi: Learning cooperative human-object interaction with manipulated object dynamics},
  author={Gao, Jiawei and Wang, Ziqin and Xiao, Zeqi and Wang, Jingbo and Wang, Tai and Cao, Jinkun and Hu, Xiaolin and Liu, Si and Dai, Jifeng and Pang, Jiangmiao},
  journal={Advances in Neural Information Processing Systems},
  volume={37},
  pages={79741--79763},
  year={2024}
}

@misc{xu2025parc,
    title={PARC: Physics-based Augmentation with Reinforcement Learning for Character Controllers},
    author={Michael Xu and Yi Shi and KangKang Yin and Xue Bin Peng},
    year={2025},
    eprint={2505.04002},
    archivePrefix={arXiv},
    primaryClass={cs.GR}
}

@article{luo2024omnigrasp,
  title={Omnigrasp: Grasping diverse objects with simulated humanoids},
  author={Luo, Zhengyi and Cao, Jinkun and Christen, Sammy and Winkler, Alexander and Kitani, Kris and Xu, Weipeng},
  journal={Advances in Neural Information Processing Systems},
  volume={37},
  pages={2161--2184},
  year={2024}
}

@inproceedings{braun2024physically,
  author = { Braun, Jona and Christen, Sammy and Kocabas, Muhammed and Aksan, Emre and Hilliges, Otmar },
booktitle = { 2024 International Conference on 3D Vision (3DV) },
title = {{ Physically Plausible Full-Body Hand-Object Interaction Synthesis }},
year = {2024},
volume = {},
ISSN = {},
pages = {464-473},
doi = {10.1109/3DV62453.2024.00109},
url = {https://doi.ieeecomputersociety.org/10.1109/3DV62453.2024.00109},
publisher = {IEEE Computer Society},
address = {Los Alamitos, CA, USA},
month =mar
}

@inproceedings{xiao2023unified,
  author       = {Zeqi Xiao and
                  Tai Wang and
                  Jingbo Wang and
                  Jinkun Cao and
                  Wenwei Zhang and
                  Bo Dai and
                  Dahua Lin and
                  Jiangmiao Pang},
  title        = {Unified Human-Scene Interaction via Prompted Chain-of-Contacts},
  booktitle    = {The Twelfth International Conference on Learning Representations,
                  {ICLR} 2024, Vienna, Austria, May 7-11, 2024},
  publisher    = {OpenReview.net},
  year         = {2024},
  url          = {https://openreview.net/forum?id=1vCnDyQkjg},
  bibsource    = {dblp computer science bibliography, https://dblp.org}
}

@article{luo2022embodied,
  title={Embodied scene-aware human pose estimation},
  author={Luo, Zhengyi and Iwase, Shun and Yuan, Ye and Kitani, Kris},
  journal={Advances in Neural Information Processing Systems},
  volume={35},
  pages={6815--6828},
  year={2022}
}

@inproceedings{lee2023questenvsim,author = {Lee, Sunmin and Starke, Sebastian and Ye, Yuting and Won, Jungdam and Winkler, Alexander},
title = {QuestEnvSim: Environment-Aware Simulated Motion Tracking from Sparse Sensors},
year = {2023},
isbn = {9798400701597},
publisher = {Association for Computing Machinery},
address = {New York, NY, USA},
url = {https://doi.org/10.1145/3588432.3591504},
doi = {10.1145/3588432.3591504},
booktitle = {ACM SIGGRAPH 2023 Conference Proceedings},
articleno = {62},
numpages = {9},
location = {Los Angeles, CA, USA},
series = {SIGGRAPH '23}
}

@inproceedings{hassan2023synthesizing,
  author = {Hassan, Mohamed and Guo, Yunrong and Wang, Tingwu and Black, Michael and Fidler, Sanja and Peng, Xue Bin},
title = {Synthesizing Physical Character-Scene Interactions},
year = {2023},
isbn = {9798400701597},
publisher = {Association for Computing Machinery},
address = {New York, NY, USA},
url = {https://doi.org/10.1145/3588432.3591525},
doi = {10.1145/3588432.3591525},
booktitle = {ACM SIGGRAPH 2023 Conference Proceedings},
articleno = {63},
numpages = {9},
location = {Los Angeles, CA, USA},
series = {SIGGRAPH '23}
}

@inproceedings{dou2022case,author = {Dou, Zhiyang and Chen, Xuelin and Fan, Qingnan and Komura, Taku and Wang, Wenping},
title = {C·ASE: Learning Conditional Adversarial Skill Embeddings for Physics-based Characters},
year = {2023},
isbn = {9798400703157},
publisher = {Association for Computing Machinery},
address = {New York, NY, USA},
url = {https://doi.org/10.1145/3610548.3618205},
doi = {10.1145/3610548.3618205},
booktitle = {SIGGRAPH Asia 2023 Conference Papers},
articleno = {2},
numpages = {11},
location = {Sydney, NSW, Australia},
series = {SA '23}
}

@inproceedings{juravsky2024superpadl,author = {Juravsky, Jordan and Guo, Yunrong and Fidler, Sanja and Peng, Xue Bin},
title = {SuperPADL: Scaling Language-Directed Physics-Based Control with Progressive Supervised Distillation},
year = {2024},
isbn = {9798400705250},
publisher = {Association for Computing Machinery},
address = {New York, NY, USA},
url = {https://doi.org/10.1145/3641519.3657492},
doi = {10.1145/3641519.3657492},
booktitle = {ACM SIGGRAPH 2024 Conference Papers},
articleno = {121},
numpages = {11},
location = {Denver, CO, USA},
series = {SIGGRAPH '24}
}

@article{
        zhu2023NCP,
        author = {Zhu, Qingxu and Zhang, He and Lan, Mengting and Han, Lei},
        title = {Neural Categorical Priors for Physics-Based Character Control},
        year = {2023},
        issue_date = {December 2023},
        publisher = {Association for Computing Machinery},
        address = {New York, NY, USA},
        volume = {42},
        number = {6},
        issn = {0730-0301},
        url = {https://doi.org/10.1145/3618397},
        doi = {10.1145/3618397},
        journal = {ACM Trans. Graph.},
        month = {dec},
        articleno = {178},
        numpages = {16}
}

@inproceedings{ren2023insactor,
author = {Ren, Jiawei and Zhang, Mingyuan and Yu, Cunjun and Ma, Xiao and Pan, Liang and Liu, Ziwei},
title = {InsActor: instruction-driven physics-based characters},
year = {2023},
publisher = {Curran Associates Inc.},
address = {Red Hook, NY, USA},
booktitle = {Proceedings of the 37th International Conference on Neural Information Processing Systems},
articleno = {2618},
numpages = {13},
location = {New Orleans, LA, USA},
series = {NIPS '23}
}

@misc{kumar2023words,
      title={Words into Action: Learning Diverse Humanoid Robot Behaviors using Language Guided Iterative Motion Refinement}, 
      author={K. Niranjan Kumar and Irfan Essa and Sehoon Ha},
      year={2023},
      eprint={2310.06226},
      archivePrefix={arXiv},
      primaryClass={cs.RO}
}

@inproceedings{
	tessler2023calm,author = {Tessler, Chen and Kasten, Yoni and Guo, Yunrong and Mannor, Shie and Chechik, Gal and Peng, Xue Bin},
title = {CALM: Conditional Adversarial Latent Models  for Directable Virtual Characters},
year = {2023},
isbn = {9798400701597},
publisher = {Association for Computing Machinery},
address = {New York, NY, USA},
url = {https://doi.org/10.1145/3588432.3591541},
doi = {10.1145/3588432.3591541},
articleno = {37},
numpages = {9},
booktitle = {ACM SIGGRAPH 2023 Conference Papers},
location = {Los Angeles, CA, USA},
series = {SIGGRAPH '23}
}

@inproceedings{winkler2022questsim,
author = {Winkler, Alexander and Won, Jungdam and Ye, Yuting},
title = {QuestSim: Human Motion Tracking from Sparse Sensors with Simulated Avatars},
year = {2022},
isbn = {9781450394703},
publisher = {Association for Computing Machinery},
address = {New York, NY, USA},
url = {https://doi.org/10.1145/3550469.3555411},
doi = {10.1145/3550469.3555411},
booktitle = {SIGGRAPH Asia 2022 Conference Papers},
articleno = {2},
numpages = {8},
location = {Daegu, Republic of Korea},
series = {SA '22}
}

@article{yao2022controlvae,
  title={Controlvae: Model-based learning of generative controllers for physics-based characters},
  author={Yao, Heyuan and Song, Zhenhua and Chen, Baoquan and Liu, Libin},
  journal={ACM Transactions on Graphics (TOG)},
  volume={41},
  number={6},
  pages={1--16},
  year={2022},
  publisher={ACM New York, NY, USA}
}

@inproceedings{
luo2024universal,author       = {Zhengyi Luo and
                  Jinkun Cao and
                  Josh Merel and
                  Alexander Winkler and
                  Jing Huang and
                  Kris M. Kitani and
                  Weipeng Xu},
  title        = {Universal Humanoid Motion Representations for Physics-Based Control},
  booktitle    = {The Twelfth International Conference on Learning Representations,
                  {ICLR} 2024, Vienna, Austria, May 7-11, 2024},
  publisher    = {OpenReview.net},
  year         = {2024},
  url          = {https://openreview.net/forum?id=OrOd8PxOO2},
  bibsource    = {dblp computer science bibliography, https://dblp.org}
}

@article{yao2024moconvq,
  title={Moconvq: Unified physics-based motion control via scalable discrete representations},
  author={Yao, Heyuan and Song, Zhenhua and Zhou, Yuyang and Ao, Tenglong and Chen, Baoquan and Liu, Libin},
  journal={ACM Transactions on Graphics (TOG)},
  volume={43},
  number={4},
  pages={1--21},
  year={2024},
  publisher={ACM New York, NY, USA}
}

@article{tessler2024maskedmimic,
  title={Maskedmimic: Unified physics-based character control through masked motion inpainting},
  author={Tessler, Chen and Guo, Yunrong and Nabati, Ofir and Chechik, Gal and Peng, Xue Bin},
  journal={ACM Transactions on Graphics (TOG)},
  volume={43},
  number={6},
  pages={1--21},
  year={2024},
  publisher={ACM New York, NY, USA}
}

@article{holt1991predicting,
  title={Predicting the minimal energy costs of human walking.},
  author={Holt, Kenneth G and Hamill, Joseph and Andres, Robert O},
  journal={Medicine and science in sports and exercise},
  volume={23},
  number={4},
  pages={491--498},
  year={1991}
}

@article{rathkey2017people,
  title={People choose to run at their optimal speed},
  author={Rathkey, Joseph K and Wall-Scheffler, Cara M},
  journal={American Journal of Physical Anthropology},
  volume={163},
  number={1},
  pages={85--93},
  year={2017},
  publisher={Wiley Online Library}
}

@article{willcockson2012reconsidering,
  title={Reconsidering the effects of respiratory constraints on the optimal running speed},
  author={Willcockson, Michael A and Wall-Scheffler, Cara M},
  journal={Medicine \& Science in Sports \& Exercise},
  volume={44},
  number={7},
  pages={1344--1350},
  year={2012},
  publisher={LWW}
}

@article{cher2015minimum,
  title={Minimum cost of transport in human running is not ubiquitous},
  author={Cher, Pei Hua and Stewart, Ian and Worringham, Charles},
  journal={Medicine and Science in Sports and Exercise},
  volume={47},
  number={2},
  pages={307--314},
  year={2015},
  publisher={Lippincott Williams \& Wilkins}
}

@article{diedrich1995change,
  title={Why change gaits? Dynamics of the walk-run transition.},
  author={Diedrich, Frederick J and Warren Jr, William H},
  journal={Journal of Experimental Psychology: Human Perception and Performance},
  volume={21},
  number={1},
  pages={183},
  year={1995},
  publisher={American Psychological Association}
}

@article{hreljac1993preferred,
  title={Preferred and energetically optimal gait transition speeds in human locomotion.},
  author={Hreljac, ALAN},
  journal={Medicine and science in sports and exercise},
  volume={25},
  number={10},
  pages={1158--1162},
  year={1993}
}

@article{steudel2009optimal,
  title={Optimal running speed and the evolution of hominin hunting strategies},
  author={Steudel-Numbers, Karen L and Wall-Scheffler, Cara M},
  journal={Journal of Human Evolution},
  volume={56},
  number={4},
  pages={355--360},
  year={2009},
  publisher={Elsevier}
}

@article{long2013walking,
  title={Walking, running, and resting under time, distance, and average speed constraints: optimality of walk--run--rest mixtures},
  author={Long III, Leroy L and Srinivasan, Manoj},
  journal={Journal of The Royal Society Interface},
  volume={10},
  number={81},
  pages={20120980},
  year={2013},
  publisher={The Royal Society}
}

@article{xia2008theoretical,
  title={A theoretical approach for modeling peripheral muscle fatigue and recovery},
  author={Xia, Ting and Law, Laura A Frey},
  journal={Journal of biomechanics},
  volume={41},
  number={14},
  pages={3046--3052},
  year={2008},
  publisher={Elsevier}
}

@inproceedings{cheema2023discovering,
  author = {Cheema, Noshaba and Xu, Rui and Kim, Nam Hee and H\"{a}m\"{a}l\"{a}inen, Perttu and Golyanik, Vladislav and Habermann, Marc and Theobalt, Christian and Slusallek, Philipp},
title = {Discovering Fatigued Movements for Virtual Character Animation},
year = {2023},
isbn = {9798400703157},
publisher = {Association for Computing Machinery},
address = {New York, NY, USA},
url = {https://doi.org/10.1145/3610548.3618176},
doi = {10.1145/3610548.3618176},
booktitle = {SIGGRAPH Asia 2023 Conference Papers},
articleno = {47},
numpages = {12},
location = {Sydney, NSW, Australia},
series = {SA '23}
}

@inproceedings{kider2011data,author = {Kider, Joseph T. and Pollock, Kaitlin and Safonova, Alla},
title = {A data-driven appearance model for human fatigue},
year = {2011},
isbn = {9781450309233},
publisher = {Association for Computing Machinery},
address = {New York, NY, USA},
url = {https://doi.org/10.1145/2019406.2019423},
doi = {10.1145/2019406.2019423},
booktitle = {Proceedings of the 2011 ACM SIGGRAPH/Eurographics Symposium on Computer Animation},
pages = {119–128},
numpages = {10},
location = {Vancouver, British Columbia, Canada},
series = {SCA '11}
}

@INPROCEEDINGS{cheng2024expressive,
  AUTHOR    = {Xuxin Cheng AND Yandong Ji AND Junming Chen AND Ruihan Yang AND Ge Yang AND Xiaolong Wang}, 
    TITLE     = {{Expressive Whole-Body Control for Humanoid Robots}}, 
    BOOKTITLE = {Proceedings of Robotics: Science and Systems}, 
    YEAR      = {2024}, 
    ADDRESS   = {Delft, Netherlands}, 
    MONTH     = {July}, 
    DOI       = {10.15607/RSS.2024.XX.107} 
}

@misc{he2025asap,
      title={ASAP: Aligning Simulation and Real-World Physics for Learning Agile Humanoid Whole-Body Skills}, 
      author={Tairan He and Jiawei Gao and Wenli Xiao and Yuanhang Zhang and Zi Wang and Jiashun Wang and Zhengyi Luo and Guanqi He and Nikhil Sobanbab and Chaoyi Pan and Zeji Yi and Guannan Qu and Kris Kitani and Jessica Hodgins and Linxi "Jim" Fan and Yuke Zhu and Changliu Liu and Guanya Shi},
      year={2025},
      eprint={2502.01143},
      archivePrefix={arXiv},
      primaryClass={cs.RO},
      url={https://arxiv.org/abs/2502.01143}, 
}

@misc{he2024hover,
      title={HOVER: Versatile Neural Whole-Body Controller for Humanoid Robots}, 
      author={Tairan He and Wenli Xiao and Toru Lin and Zhengyi Luo and Zhenjia Xu and Zhenyu Jiang and Jan Kautz and Changliu Liu and Guanya Shi and Xiaolong Wang and Linxi Fan and Yuke Zhu},
      year={2025},
      eprint={2410.21229},
      archivePrefix={arXiv},
      primaryClass={cs.RO},
      url={https://arxiv.org/abs/2410.21229}, 
}

@inproceedings{cheema2020predicting,
  author = {Cheema, Noshaba and Frey-Law, Laura A. and Naderi, Kourosh and Lehtinen, Jaakko and Slusallek, Philipp and H\"{a}m\"{a}l\"{a}inen, Perttu},
title = {Predicting Mid-Air Interaction Movements and Fatigue Using Deep Reinforcement Learning},
year = {2020},
isbn = {9781450367080},
publisher = {Association for Computing Machinery},
address = {New York, NY, USA},
url = {https://doi.org/10.1145/3313831.3376701},
doi = {10.1145/3313831.3376701},
booktitle = {Proceedings of the 2020 CHI Conference on Human Factors in Computing Systems},
pages = {1–13},
numpages = {13},
location = {Honolulu, HI, USA},
series = {CHI '20}
}

@inproceedings{feng2023musclevae,
  author = {Feng, Yusen and Xu, Xiyan and Liu, Libin},
title = {MuscleVAE: Model-Based Controllers of Muscle-Actuated Characters},
year = {2023},
isbn = {9798400703157},
publisher = {Association for Computing Machinery},
address = {New York, NY, USA},
url = {https://doi.org/10.1145/3610548.3618137},
doi = {10.1145/3610548.3618137},
booktitle = {SIGGRAPH Asia 2023 Conference Papers},
articleno = {3},
numpages = {11},
location = {Sydney, NSW, Australia},
series = {SA '23}
}

@article{komura2000creating,
  title={Creating and retargetting motion by the musculoskeletal human body model},
  author={Komura, Taku and Shinagawa, Yoshihisa and Kunii, Tosiyasu L},
  journal={The visual computer},
  volume={16},
  pages={254--270},
  year={2000},
  publisher={Springer}
}

@article{luo2021dynamics,
  title={Dynamics-regulated kinematic policy for egocentric pose estimation},
  author={Luo, Zhengyi and Hachiuma, Ryo and Yuan, Ye and Kitani, Kris},
  journal={Advances in Neural Information Processing Systems},
  volume={34},
  pages={25019--25032},
  year={2021}
}

@inproceedings{mahmood2019amass,
author = {N. Mahmood and N. Ghorbani and N. F. Troje and G. Pons-Moll and M. Black},
booktitle = {2019 IEEE/CVF International Conference on Computer Vision (ICCV)},
title = {AMASS: Archive of Motion Capture As Surface Shapes},
year = {2019},
volume = {},
issn = {},
pages = {5441-5450},
doi = {10.1109/ICCV.2019.00554},
url = {https://doi.ieeecomputersociety.org/10.1109/ICCV.2019.00554},
publisher = {IEEE Computer Society},
address = {Los Alamitos, CA, USA},
month = {nov}
}

@inproceedings{luo2023perpetual,
author = {Z. Luo and J. Cao and A. Winkler and K. Kitani and W. Xu},
booktitle = {2023 IEEE/CVF International Conference on Computer Vision (ICCV)},
title = {Perpetual Humanoid Control for Real-time Simulated Avatars},
year = {2023},
volume = {},
issn = {},
pages = {10861-10870},
doi = {10.1109/ICCV51070.2023.01000},
url = {https://doi.ieeecomputersociety.org/10.1109/ICCV51070.2023.01000},
publisher = {IEEE Computer Society},
address = {Los Alamitos, CA, USA},
month = {oct}
}

@inproceedings{truong2024pdp,
author = {Truong, Takara Everest and Piseno, Michael and Xie, Zhaoming and Liu, Karen},
title = {PDP: Physics-Based Character Animation via Diffusion Policy},
year = {2024},
isbn = {9798400711312},
publisher = {Association for Computing Machinery},
address = {New York, NY, USA},
url = {https://doi.org/10.1145/3680528.3687683},
doi = {10.1145/3680528.3687683},
booktitle = {SIGGRAPH Asia 2024 Conference Papers},
articleno = {86},
numpages = {10},
location = {Tokyo, Japan},
series = {SA '24}
}

@article{chi2023diffusion,
author = {Cheng Chi and Zhenjia Xu and Siyuan Feng and Eric Cousineau and Yilun Du and Benjamin Burchfiel and Russ Tedrake and Shuran Song},
title ={Diffusion policy: Visuomotor policy learning via action diffusion},
journal = {The International Journal of Robotics Research},
volume = {0},
number = {0},
pages = {02783649241273668},
year = {0},
doi = {10.1177/02783649241273668},
URL = {https://doi.org/10.1177/02783649241273668},
eprint = {https://doi.org/10.1177/02783649241273668}
}

@article{peng2022ase,author = {Peng, Xue Bin and Guo, Yunrong and Halper, Lina and Levine, Sergey and Fidler, Sanja},
title = {ASE: large-scale reusable adversarial skill embeddings for physically simulated characters},
year = {2022},
issue_date = {July 2022},
publisher = {Association for Computing Machinery},
address = {New York, NY, USA},
volume = {41},
number = {4},
issn = {0730-0301},
url = {https://doi.org/10.1145/3528223.3530110},
doi = {10.1145/3528223.3530110},
journal = {ACM Trans. Graph.},
month = jul,
articleno = {94},
numpages = {17}
}

@article{xu2023composite,
  title={Composite motion learning with task control},
  author={Xu, Pei and Shang, Xiumin and Zordan, Victor and Karamouzas, Ioannis},
  journal={ACM Transactions on Graphics (TOG)},
  volume={42},
  number={4},
  pages={1--16},
  year={2023},
  publisher={ACM New York, NY, USA}
}

@article{molen1972graphic,
  title={Graphic representation of the relationship between oxygen-consumption and characteristics of normal gait of the human male.},
  author={Molen, NH and Rozendal, RH and Boon, W},
  journal={Proceedings of the Koninklijke Nederlandse Akademie Van Wetenschappen. Series C. Biological and Medical Sciences},
  volume={75},
  number={4},
  pages={305--314},
  year={1972}
}

@inproceedings{loper2023smpl,
author = {Loper, Matthew and Mahmood, Naureen and Romero, Javier and Pons-Moll, Gerard and Black, Michael J.},
title = {SMPL: a skinned multi-person linear model},
year = {2015},
issue_date = {November 2015},
publisher = {Association for Computing Machinery},
address = {New York, NY, USA},
volume = {34},
issn = {0730-0301},
url = {https://doi.org/10.1145/2816795.2818013},
doi = {10.1145/2816795.2818013},
journal = {ACM Trans. Graph.},
booktitle = {ACM Transactions on Graphics},
month = {oct},
articleno = {248},
numpages = {16}
}

@inproceedings{araujo2023circle,
author = { Araujo, Joao Pedro and Li, Jiaman and Vetrivel, Karthik and Agarwal, Rishi and Wu, Jiajun and Gopinath, Deepak and Clegg, Alexander and Liu, C. Karen },
booktitle = { 2023 IEEE/CVF Conference on Computer Vision and Pattern Recognition (CVPR) },
title = {{ CIRCLE: Capture In Rich Contextual Environments }},
year = {2023},
volume = {},
ISSN = {},
pages = {21211-21221},
doi = {10.1109/CVPR52729.2023.02032},
url = {https://doi.ieeecomputersociety.org/10.1109/CVPR52729.2023.02032},
publisher = {IEEE Computer Society},
address = {Los Alamitos, CA, USA},
month =Jun
}

@inproceedings{makoviychuk2021isaac,
  author       = {Viktor Makoviychuk and
                  Lukasz Wawrzyniak and
                  Yunrong Guo and
                  Michelle Lu and
                  Kier Storey and
                  Miles Macklin and
                  David Hoeller and
                  Nikita Rudin and
                  Arthur Allshire and
                  Ankur Handa and
                  Gavriel State},
  editor       = {Joaquin Vanschoren and
                  Sai{-}Kit Yeung},
  title        = {Isaac Gym: High Performance {GPU} Based Physics Simulation For Robot Learning},
  booktitle    = {Proceedings of the Neural Information Processing Systems Track on Datasets and Benchmarks},
  year         = {2021},
  url          = {https://datasets-benchmarks-proceedings.neurips.cc/paper/2021/hash/28dd2c7955ce926456240b2ff0100bde-Abstract-round2.html},
  bibsource    = {dblp computer science bibliography, https://dblp.org},
address = {Virtual},
publisher = {Curran Associates Inc.},
}

@article{arellano2011effects,
  title={The effects of step width and arm swing on energetic cost and lateral balance during running},
  author={Arellano, Christopher J and Kram, Rodger},
  journal={Journal of biomechanics},
  volume={44},
  number={7},
  pages={1291--1295},
  year={2011},
  publisher={Elsevier}
}

@book{alexander1996optima,
 ISBN = {9780691027982},
 URL = {http://www.jstor.org/stable/j.ctv173f0gj},
 author = {R. McNeill Alexander},
 publisher = {Princeton University Press},
 title = {Optima for Animals: Revised Edition},
 urldate = {2025-05-23},
 year = {1996}
}

@article{elftman1966biomechanics,
  title={Biomechanics of muscle: with particular application to studies of gait},
  author={Elftman, Herbert},
  journal={JBJS},
  volume={48},
  number={2},
  pages={363--377},
  year={1966},
  publisher={LWW}
}

@article{atzler1927arbeitsphysiologische,
  title={Arbeitsphysiologische studien: Iii. teil},
  author={Atzler, Edgar and Herbst, Robert},
  journal={Pfl{\"u}ger's Archiv f{\"u}r die gesamte Physiologie des Menschen und der Tiere},
  volume={215},
  pages={291--328},
  year={1927},
  publisher={Springer}
}

@article{gottschall2003energy,
  title={Energy cost and muscular activity required for propulsion during walking},
  author={Gottschall, Jinger S and Kram, Rodger},
  journal={Journal of Applied Physiology},
  volume={94},
  number={5},
  pages={1766--1772},
  year={2003},
  publisher={American Physiological Society Bethesda, MD}
}

@article{donelan2004mechanical,
  title={Mechanical and metabolic requirements for active lateral stabilization in human walking},
  author={Donelan, J Maxwell and Shipman, David W and Kram, Rodger and Kuo, Arthur D},
  journal={Journal of biomechanics},
  volume={37},
  number={6},
  pages={827--835},
  year={2004},
  publisher={Elsevier}
}

@article{behavioral,
    author = {McAllister, Megan J. and Chen, Anthony and Selinger, Jessica C.},
    title = {Behavioural energetics in human locomotion: how energy use influences how we move},
    journal = {Journal of Experimental Biology},
    volume = {228},
    pages = {JEB248125},
    year = {2025},
    month = {02},
    issn = {0022-0949},
    doi = {10.1242/jeb.248125},
    url = {https://doi.org/10.1242/jeb.248125},
    eprint = {https://journals.biologists.com/jeb/article-pdf/228/Suppl\_1/JEB248125/3634154/jeb248125.pdf},
}

@article{ralston1958energy,
  title={Energy-speed relation and optimal speed during level walking},
  author={Ralston, Henry J},
  journal={Internationale Zeitschrift f{\"u}r Angewandte Physiologie Einschliesslich Arbeitsphysiologie},
  volume={17},
  number={4},
  pages={277--283},
  year={1958},
  publisher={Springer}
}

@article{brooks2005basics,
  title={Basics of metabolism},
  author={Brooks, GA and Fahey, TD and White, TP and Baldwin, KM},
  journal={Exercise Physiology Human Bioenergetics and its Applications},
  volume={4},
  pages={43--58},
  year={2005}
}

@article{brown2021unified,
  title={A unified energy-optimality criterion predicts human navigation paths and speeds},
  author={Brown, Geoffrey L and Seethapathi, Nidhi and Srinivasan, Manoj},
  journal={Proceedings of the National Academy of Sciences},
  volume={118},
  number={29},
  pages={e2020327118},
  year={2021},
  publisher={National Academy of Sciences}
}

@inproceedings{ross2011reduction,
  title={A reduction of imitation learning and structured prediction to no-regret online learning},
  author={Ross, St{\'e}phane and Gordon, Geoffrey and Bagnell, Drew},
  booktitle={Proceedings of the fourteenth international conference on artificial intelligence and statistics},
  pages={627--635},
  year={2011},
  organization={JMLR Workshop and Conference Proceedings}
}

@article{daniels2023human,
  title={Human locomotion over obstacles reveals real-time prediction of energy expenditure for optimized decision-making},
  author={Daniels, Katherine AJ and Burn, JF},
  journal={Proceedings of the Royal Society B},
  volume={290},
  number={2000},
  pages={20230200},
  year={2023},
  publisher={The Royal Society}
}

@article{de2010feature,
  title={Feature-based locomotion controllers},
  author={De Lasa, Martin and Mordatch, Igor and Hertzmann, Aaron},
  journal={ACM transactions on graphics (TOG)},
  volume={29},
  number={4},
  pages={1--10},
  year={2010},
  publisher={ACM New York, NY, USA}
}

@article{geijtenbeek2013flexible,
  title={Flexible muscle-based locomotion for bipedal creatures},
  author={Geijtenbeek, Thomas and Van De Panne, Michiel and Van Der Stappen, A Frank},
  journal={ACM Transactions on Graphics (TOG)},
  volume={32},
  number={6},
  pages={1--11},
  year={2013},
  publisher={ACM New York, NY, USA}
}

@article{liu2010sampling,
author = {Liu, Libin and Yin, KangKang and van de Panne, Michiel and Shao, Tianjia and Xu, Weiwei},
title = {Sampling-based contact-rich motion control},
year = {2010},
issue_date = {July 2010},
publisher = {Association for Computing Machinery},
address = {New York, NY, USA},
volume = {29},
number = {4},
issn = {0730-0301},
url = {https://doi.org/10.1145/1778765.1778865},
doi = {10.1145/1778765.1778865},
journal = {ACM Trans. Graph.},
month = jul,
articleno = {128},
numpages = {10}
}

@article{yuan2020residual,
  title={Residual force control for agile human behavior imitation and extended motion synthesis},
  author={Yuan, Ye and Kitani, Kris},
  journal={Advances in Neural Information Processing Systems},
  volume={33},
  pages={21763--21774},
  year={2020}
}

@inproceedings{ho2016generative,
 author = {Ho, Jonathan and Ermon, Stefano},
 booktitle = {Advances in Neural Information Processing Systems},
 editor = {D. Lee and M. Sugiyama and U. Luxburg and I. Guyon and R. Garnett},
 pages = {},
 publisher = {Curran Associates, Inc.},
 title = {Generative Adversarial Imitation Learning},
 url = {https://proceedings.neurips.cc/paper_files/paper/2016/file/cc7e2b878868cbae992d1fb743995d8f-Paper.pdf},
 volume = {29},
 year = {2016}
}

@article{yu2018learning,
  title={Learning symmetric and low-energy locomotion},
  author={Yu, Wenhao and Turk, Greg and Liu, C Karen},
  journal={ACM Transactions on Graphics (TOG)},
  volume={37},
  number={4},
  pages={1--12},
  year={2018},
  publisher={ACM New York, NY, USA}
}

@article{peng2021amp,
  title={Amp: Adversarial motion priors for stylized physics-based character control},
  author={Peng, Xue Bin and Ma, Ze and Abbeel, Pieter and Levine, Sergey and Kanazawa, Angjoo},
  journal={ACM Transactions on Graphics (ToG)},
  volume={40},
  number={4},
  pages={1--20},
  year={2021},
  publisher={ACM New York, NY, USA}
}

@misc{merel2017learning,
      title={Learning human behaviors from motion capture by adversarial imitation}, 
      author={Josh Merel and Yuval Tassa and Dhruva TB and Sriram Srinivasan and Jay Lemmon and Ziyu Wang and Greg Wayne and Nicolas Heess},
      year={2017},
      eprint={1707.02201},
      archivePrefix={arXiv},
      primaryClass={cs.RO},
      url={https://arxiv.org/abs/1707.02201}, 
}

@article{peng2018deepmimic,
  title={Deepmimic: Example-guided deep reinforcement learning of physics-based character skills},
  author={Peng, Xue Bin and Abbeel, Pieter and Levine, Sergey and Van de Panne, Michiel},
  journal={ACM Transactions On Graphics (TOG)},
  volume={37},
  number={4},
  pages={1--14},
  year={2018},
  publisher={ACM New York, NY, USA}
}

@article{liu2018learning,
  title={Learning basketball dribbling skills using trajectory optimization and deep reinforcement learning},
  author={Liu, Libin and Hodgins, Jessica},
  journal={Acm transactions on graphics (tog)},
  volume={37},
  number={4},
  pages={1--14},
  year={2018},
  publisher={ACM New York, NY, USA}
}

@article{liu2017learning,
  title={Learning to schedule control fragments for physics-based characters using deep q-learning},
  author={Liu, Libin and Hodgins, Jessica},
  journal={ACM Transactions on Graphics (TOG)},
  volume={36},
  number={3},
  pages={1--14},
  year={2017},
  publisher={ACM New York, NY, USA}
}

@article{riveros2025silico,
  title={An in-silico investigation of the effect of changing cycling crank power and cadence on muscle energetics and active muscle volume},
  author={Riveros-Matthey, Cristian D and Carroll, Timothy J and Connick, Mark J and Lichtwark, Glen A},
  journal={Journal of Biomechanics},
  volume={180},
  pages={112530},
  year={2025},
  publisher={Elsevier}
}

@article{brisswalter2000energetically,
  title={Energetically optimal cadence vs. freely-chosen cadence during cycling: effect of exercise duration},
  author={Brisswalter, J and Hausswirth, C and Smith, D and Vercruyssen, F and Vallier, JM},
  journal={International journal of sports medicine},
  volume={21},
  number={01},
  pages={60--64},
  year={2000},
  publisher={Georg Thieme Verlag Stuttgart{\textperiodcentered} New York}
}

@article{bertram2001multiple,
  title={Multiple walking speed--frequency relations are predicted by constrained optimization},
  author={Bertram, John EA and Ruina, Andy},
  journal={Journal of theoretical Biology},
  volume={209},
  number={4},
  pages={445--453},
  year={2001},
  publisher={Elsevier}
}

@article{mercier1994energy,
  title={Energy expenditure and cardiorespiratory responses at the transition between walking and running},
  author={Mercier, Jacques and Gallais, Daniel Le and Durand, Marc and Goudal, Christine and Micallef, Jean Paul and Pr{\'e}faut, Christian},
  journal={European journal of applied physiology and occupational physiology},
  volume={69},
  number={6},
  pages={525--529},
  year={1994},
  publisher={Springer}
}

\appendix
\section{Additional Qualitative Results}

The following figures provide additional qualitative examples and analyses.

\begin{figure*}[t]
  \centering
  \includegraphics[width=.8\linewidth]{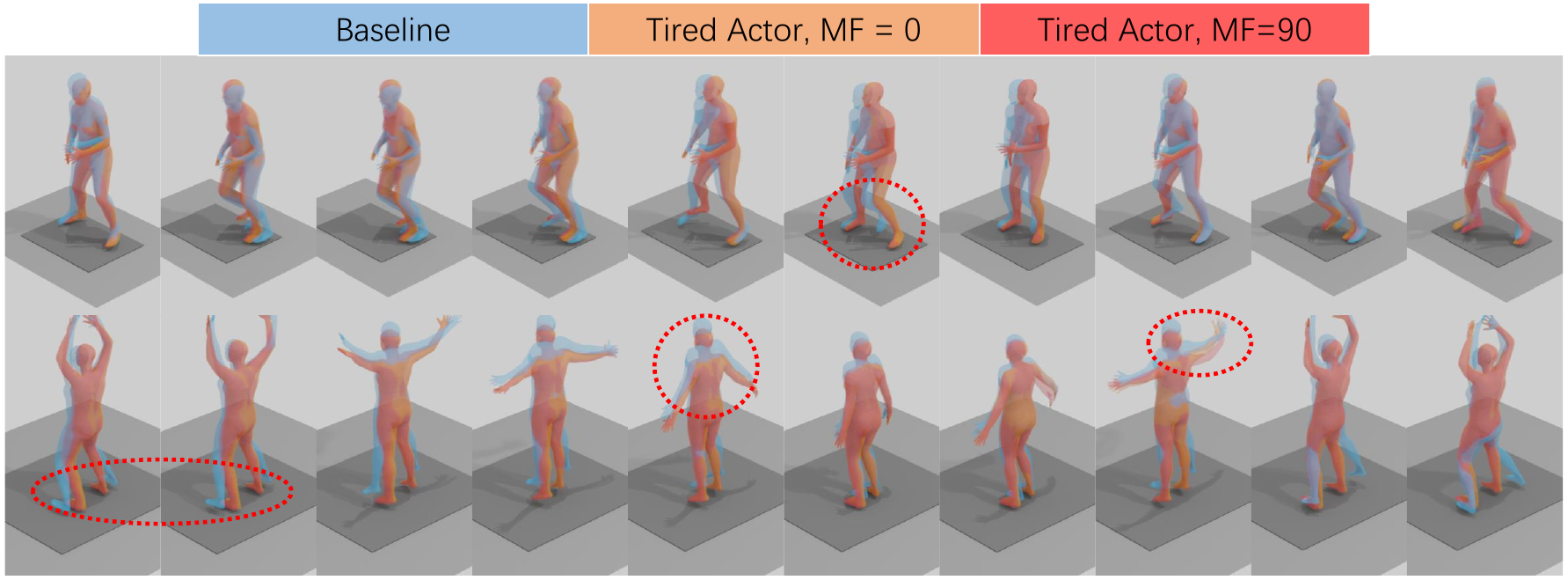}
  \caption{Motion imitation with different initial $M_F$s.}
  \label{fig:fatigue-level}
\end{figure*}

\begin{figure*}[t]
  \centering
  \includegraphics[width=.8\linewidth]{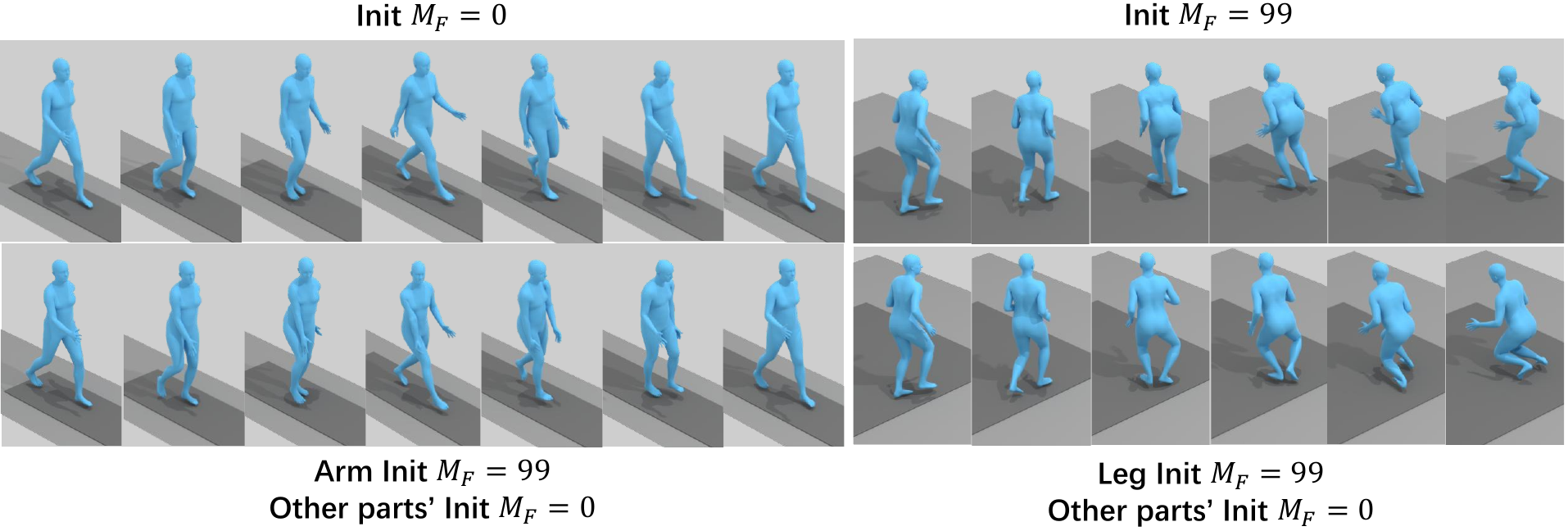}
  \caption{Fatigue-induced cross-part coordination.}
  \label{fig:different-part}
\end{figure*}

\begin{figure*}[t]
  \centering
  \includegraphics[width=.8\linewidth]{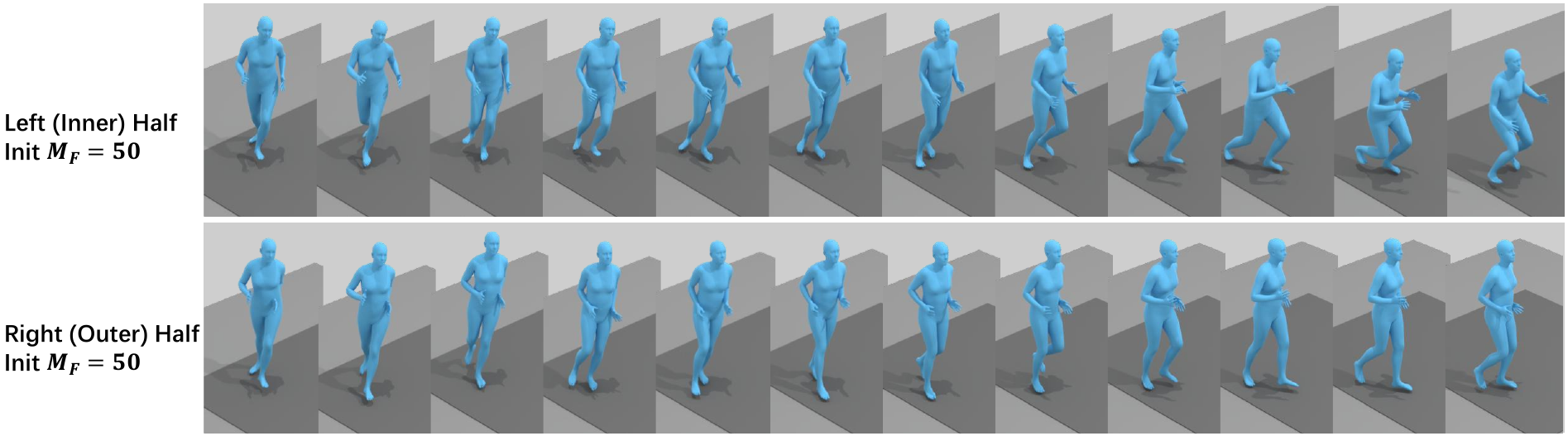}
  \caption{Fatigue-induced asymmetry. Fatigue on inner-/outer-half body results in diverse behaviors.}
  \label{fig:left-right}
\end{figure*}

\begin{figure*}[t]
  \centering
  \includegraphics[width=.8\linewidth]{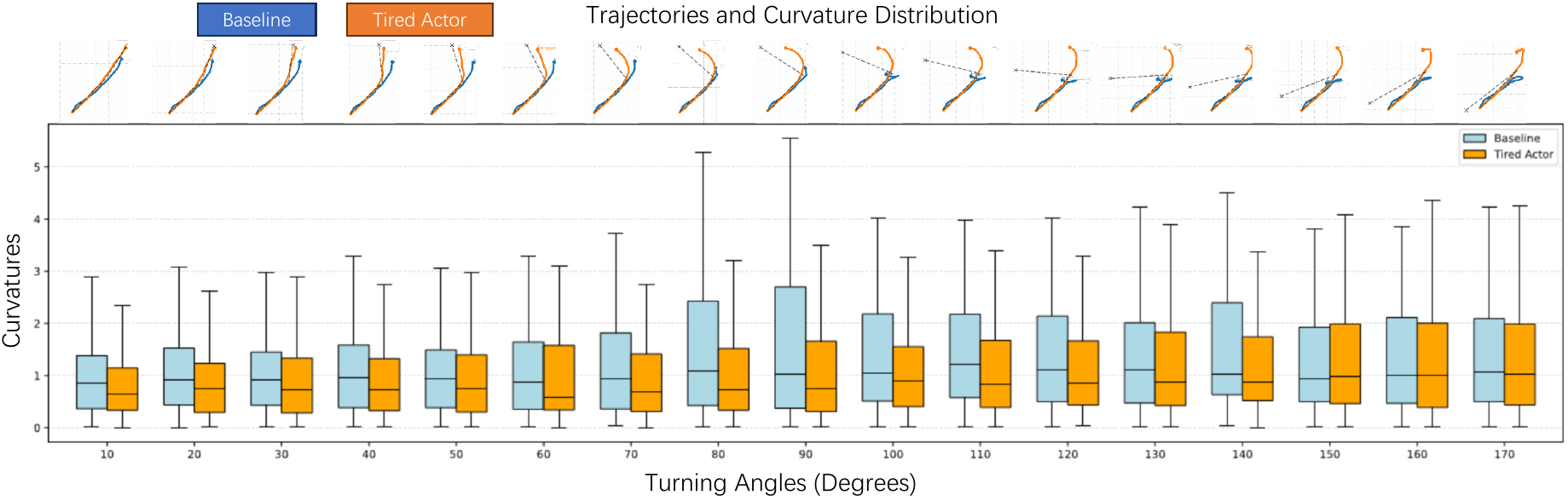}
  \caption{Fatigue-induced spatial decision making. Trajectories and curvature distributions under different turning angles are visualized. Tired Actor typically avoids sharp turns with lower curvatures, trading off target-reaching speed for sharp turns.}
  \label{fig:spatial-decision}
\end{figure*}

\begin{figure*}[t]
  \centering
  \begin{minipage}[t]{.56\linewidth}
    \centering
    \includegraphics[width=\linewidth]{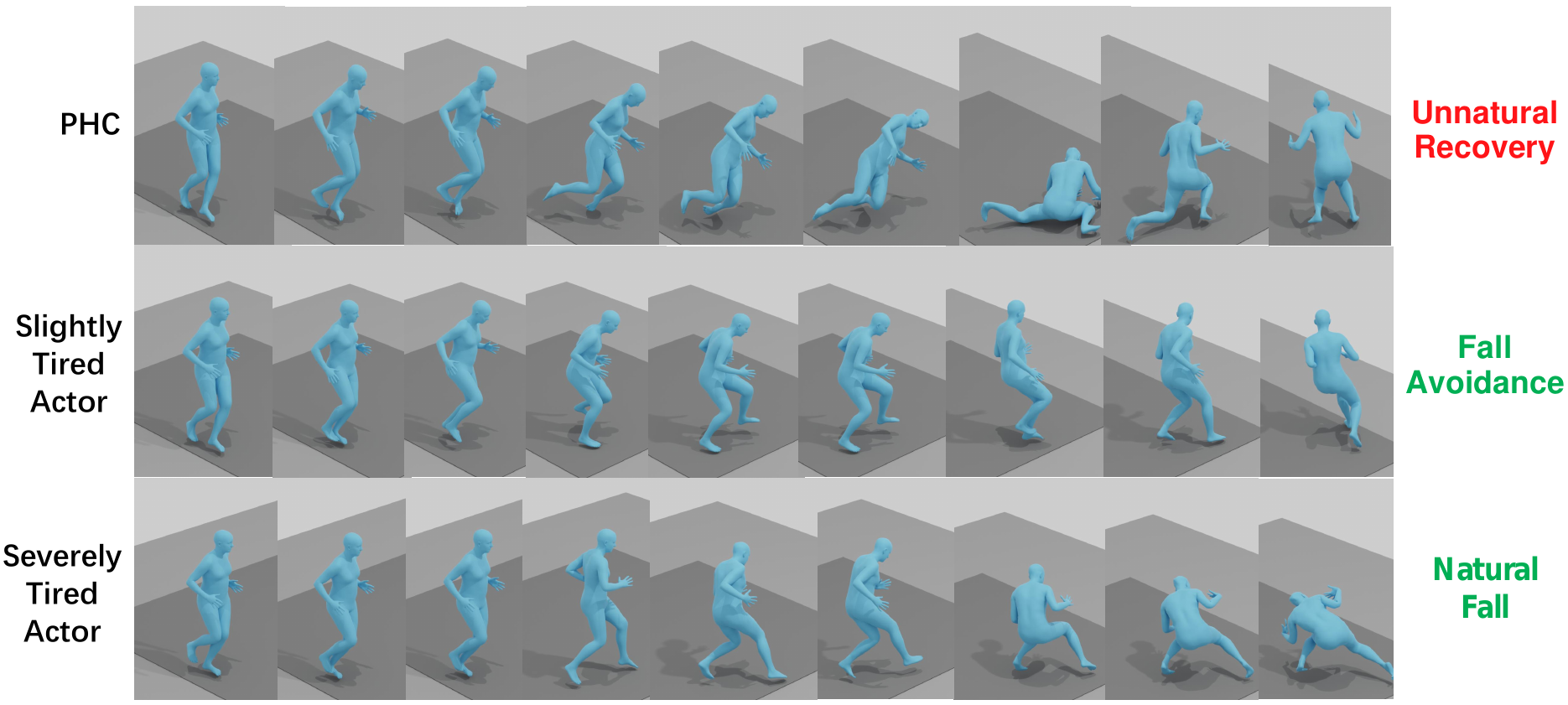}
    \caption{Fatigue-induced compensation under different $M_F$s. Though PHC manages to recover after perturbation, unnatural gestures could be observed during recovery. Instead, Tired Actor adaptively responds to external perturbations with natural behaviors.}
    \label{fig:perturbation-fatigue}
  \end{minipage}
  \hfill
  \begin{minipage}[t]{.40\linewidth}
    \centering
    \includegraphics[width=\linewidth]{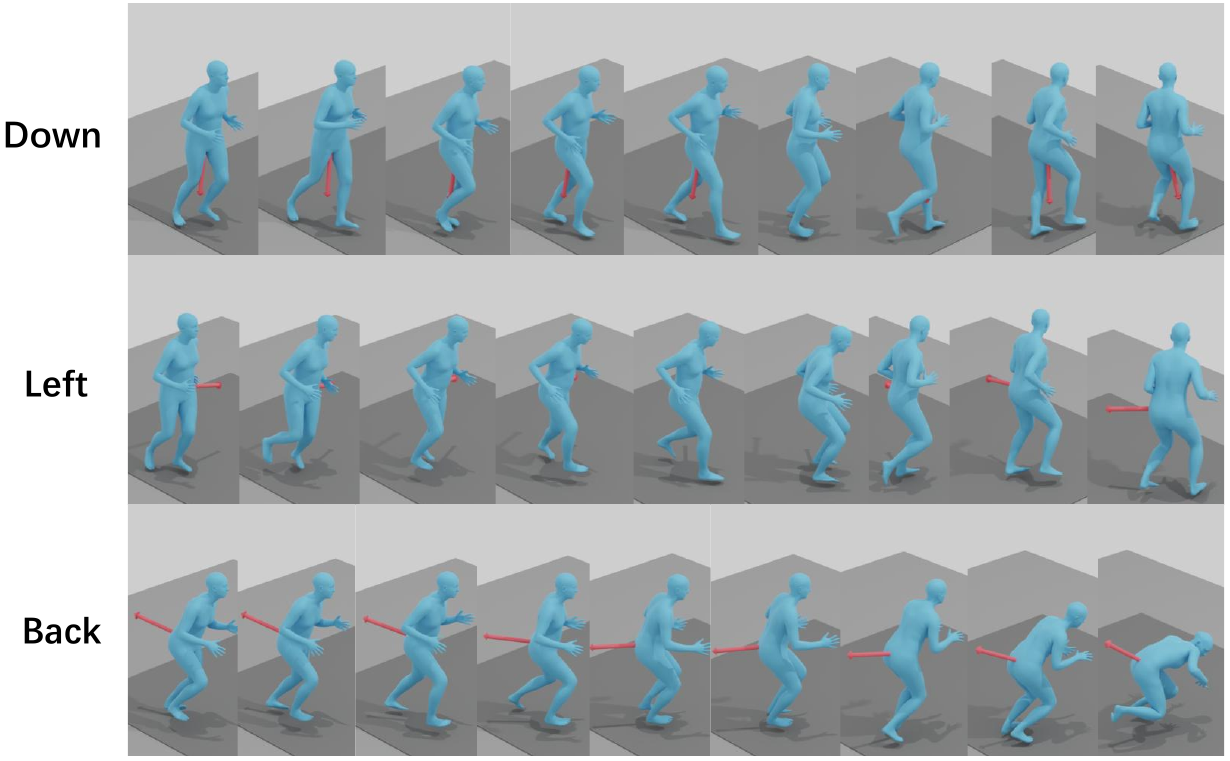}
    \caption{Fatigue-induced compensation under different directions. Lateral and downward perturbations are easier to absorb than backward perturbations.}
    \label{fig:perturbation-direction}
  \end{minipage}
\end{figure*}

\begin{figure*}[t]
  \centering
  \includegraphics[width=.45\linewidth]{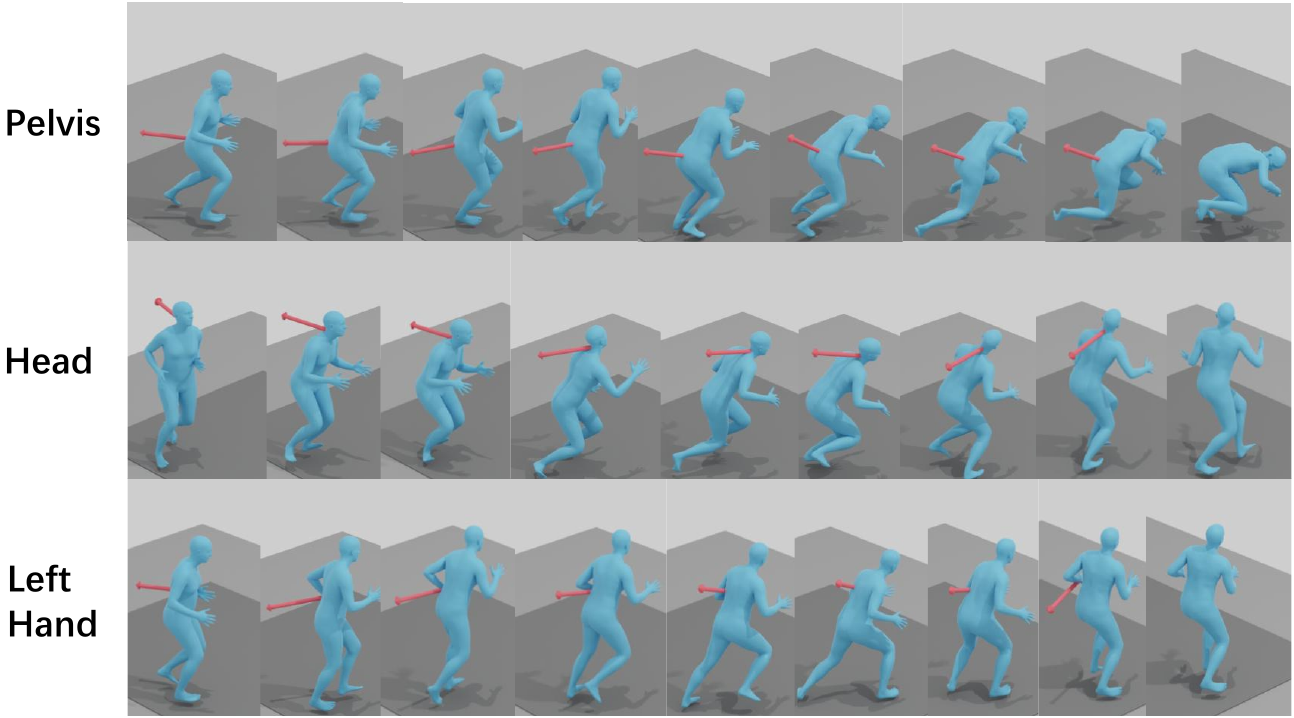}
  \caption{External perturbation compensation with different perturbation parts. Perturbations applied to distal segments are easier to handle than those applied to central body parts.}
  \label{fig:perturbation-part}
\end{figure*}

\begin{figure*}[t]
  \centering
  \includegraphics[width=.8\linewidth]{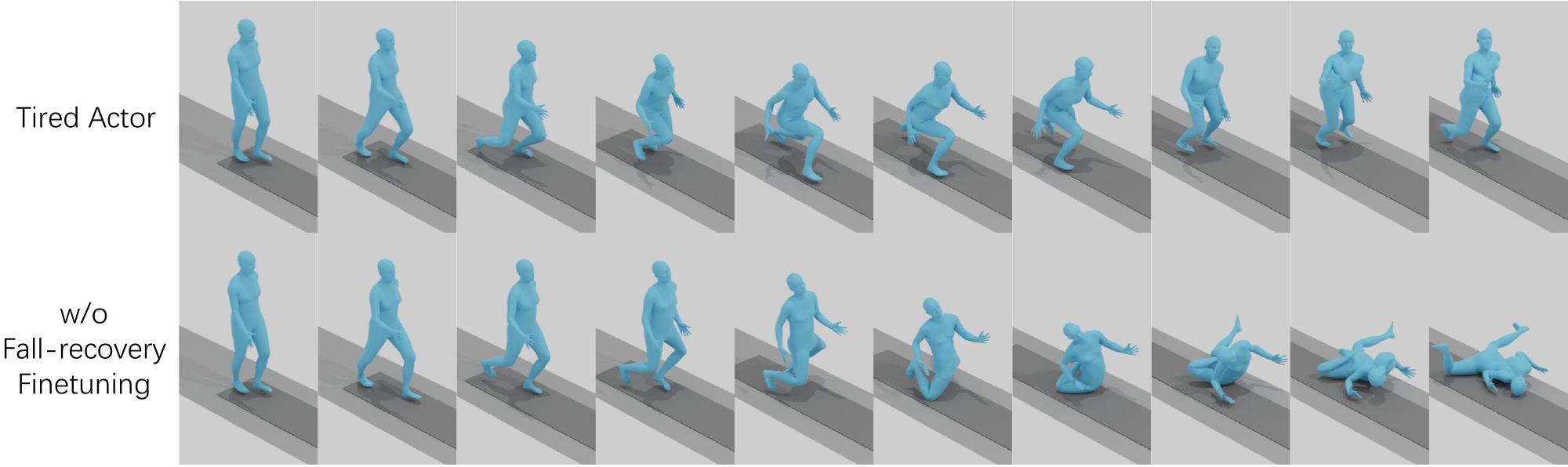}
  \caption{Tired Actors with and without fall-recovery fine-tuning imitate a forward-walking motion with initial $M_F=95\%$. The fall-recovery fine-tuning enhances the robustness of Tired Actor and prevents it from falling.}
  \label{fig:getup}
\end{figure*}

\end{document}